\documentclass[letterpaper]{article} % DO NOT CHANGE THIS
\usepackage{aaai23}  % DO NOT CHANGE THIS
\usepackage{times}  % DO NOT CHANGE THIS
\usepackage{helvet}  % DO NOT CHANGE THIS
\usepackage{courier}  % DO NOT CHANGE THIS
\usepackage[hyphens]{url}  % DO NOT CHANGE THIS
\usepackage{graphicx} % DO NOT CHANGE THIS
\usepackage{natbib}  % DO NOT CHANGE THIS AND DO NOT ADD ANY OPTIONS TO IT
\usepackage{caption} % DO NOT CHANGE THIS AND DO NOT ADD ANY OPTIONS TO IT
\usepackage{algorithm}
\usepackage{algorithmic}

\usepackage{newfloat}
\usepackage{listings}
\DeclareCaptionStyle{ruled}{labelfont=normalfont,labelsep=colon,strut=off} % DO NOT CHANGE THIS
\floatstyle{ruled}
\newfloat{listing}{tb}{lst}{}
\floatname{listing}{Listing}
\nocopyright %-- Your paper will not be published if you use this command
\author{Fabio Brau, Giulio Rossolini, Alessandro Biondi and Giorgio Buttazzo
    \thanks{\texttt{email: name.surname@santannapisa.it}\\ Under review.}}
\affiliations{%
    Department of Excellence in Robotics and AI, Scuola Superiore Sant’Anna, Pisa, Italy\\
}

\usepackage[inline]{enumitem}
\usepackage{amsthm}
\usepackage{amsmath}
\usepackage{graphicx}
\usepackage{bbm}
\usepackage{amssymb}
\usepackage{xcolor}
\usepackage{caption}
\usepackage{todonotes}
\usepackage{pifont}% http://ctan.org/pkg/pifont
\usepackage{booktabs}
\usepackage{bm}
\usepackage{subcaption} 
\newcommand{\add}[1]{{\color{black} #1}}

\newtheorem{thm}{Theorem}
\newtheorem{lem}{Lemma}

\theoremstyle{definition}
\newtheorem{defn}{Definition}

\newtheorem{oss}{Observation}
\newcommand{\BB}{\mathcal{B}}

\newcommand{\FF}{\mathcal{F}}

\newcommand{\LL}{\mathcal{L}}
\newcommand{\MM}{\mathcal{M}}
\newcommand{\UU}{\mathcal{U}}

\newcommand{\XX}{\mathcal{X}}
\newcommand{\YY}{\mathcal{Y}}
\newcommand{\KK}{\mathcal{K}}
\newcommand{\E}{\mathbb{E}}
\newcommand{\R}{\mathbb{R}}

\newcommand{\N}{\mathbb{N}}

\newcommand{\1}{\mathbbm{1}}
\newcommand{\sgn}{\mathrm{sgn}}
\newcommand{\abs}{\mathrm{abs}}
\newcommand{\oplu}{\mathrm{OPLU}}
\newcommand{\jac}{\mathrm{Jac}}
\newcommand{\diag}{\mathrm{diag}}
\newcommand{\argmax}{\mathrm{arg}\!\max}

\newcommand{\cmark}{\ding{51}}%
\newcommand{\xmark}{\ding{55}}%
\title{Robust-by-Design Classification via Unitary-Gradient Neural Networks}

\begin{document}
\maketitle
\begin{abstract}
The use of neural networks in safety-critical systems requires safe and robust models, due to the existence of adversarial attacks.
Knowing the minimal adversarial perturbation of any input $x$, or, equivalently, knowing the distance of $x$ from the classification boundary, allows evaluating the classification robustness, providing certifiable predictions. Unfortunately, state-of-the-art techniques for computing such a distance are computationally expensive
%, since involve greedy algorithms, and
and hence not suited for online applications. 
This work proposes a novel family of classifiers, namely \emph{Signed Distance Classifiers} (SDCs), that, from a theoretical perspective, directly output the exact distance of $x$ from the classification boundary, rather than a probability score (e.g., SoftMax). 
SDCs represent a family of robust-by-design classifiers.
To practically address the theoretical requirements of a SDC, a novel network architecture named \emph{Unitary-Gradient Neural Network} is presented.
Experimental results show that the proposed architecture approximates a signed distance classifier, hence allowing an online certifiable classification of $x$ at the cost of a single inference.
%We believe that this work may represent an important milestone for safe and trustworthy neural networks in safety critical systems.
\end{abstract}

% G - Direi di estendere l'abstract sulla parte di esperimenti...
% To practically address the theoretical requirements of a signed distance classifier, we present a novel architecture, namely \textit{Unitary Gradient Neural Network}. 
% That network represents, to the best of our records, the first implementation of a signed distance classifier, allowing an online certifiable prediction of $x$ at the cost of a single forward. 
%Its benefits are assessed through several comparisons with 1-Lipschitz neural networks on the CIFAR-10 dataset, resulting in a state-of-the-art online lower bound of the boundary distance and a comparable classification accuracy. 
% In the face of these contributions, we believe that this work may represent an important milestone for safe and trustworthy neural networks (sentiti libero di aggiungere "in safety critical systems")

\section{Introduction}
Deep Neural Networks (DNNs) reached popularity due to the high capability of achieving super-human performance in various tasks, such as \textit{Image Classification, Object Detection} and \textit {Image Generation}.
However, their usage in safety-critical systems, such as \textit{autonomous cars}, is pushing the scientific community toward the definition and the achievement of certifiable guarantees.

In this regard, as independently shown by \cite{szegedy, biggio}, neural networks are highly sensitive to small perturbations of the input, also known as \textit{adversarial examples}, which are not easy to detect \cite{biggio_2018, Carlini_Dill_2018,rossolini2022increasing}, and cause the model to produce a wrong classification. Informally speaking, a classifier is said to be $\varepsilon$-\emph{robust} in a certain input $x$ if the classification result does not change by perturbing $x$ with all possible perturbations of a bounded magnitude $\varepsilon$.

In the last few years, a large number of methods for crafting adversarial
examples have been presented \cite{goodfellow,deepfool, rony_2020, madry}. 
In particular, \cite{carlini, rony_2019} proposed methods to find the minimal adversarial perturbation (MAP) or, equivalently, the closest adversarial example for a given input $x$. 
% --------------------------------------------------------
Such a perturbation directly provides the distance of $x$ from the classification boundary, which, given a maximum magnitude of perturbation, can be used to verify the trustworthiness of the prediction \cite{Weng_Daniel_2018a} and design robust classifiers \cite{Wong_Kolter_2018b, Cohen_Kolter_2019}. Note that, when the MAP is known, one can check on-line whether a certain input $x$ can be perturbed with a bounded-magnitude perturbation to change the classification result. If this is the case, the network itself can signal the unsafeness of the result. Unfortunately, due the hard complexity of the algorithms for solving the MAP problem on classic models, the aforementioned strategies 
%\cite{brau}, the aforementioned methods require multiple evaluations of the model, thus resulting
are not suited for efficiently certifying the robustness of classifiers~\cite{brau}.
%---------------------------------------------------------
\begin{figure}[t]
    \centering
    \includegraphics[clip, width=\columnwidth, trim=8.7cm 3.3cm 6.2cm 2.2cm]{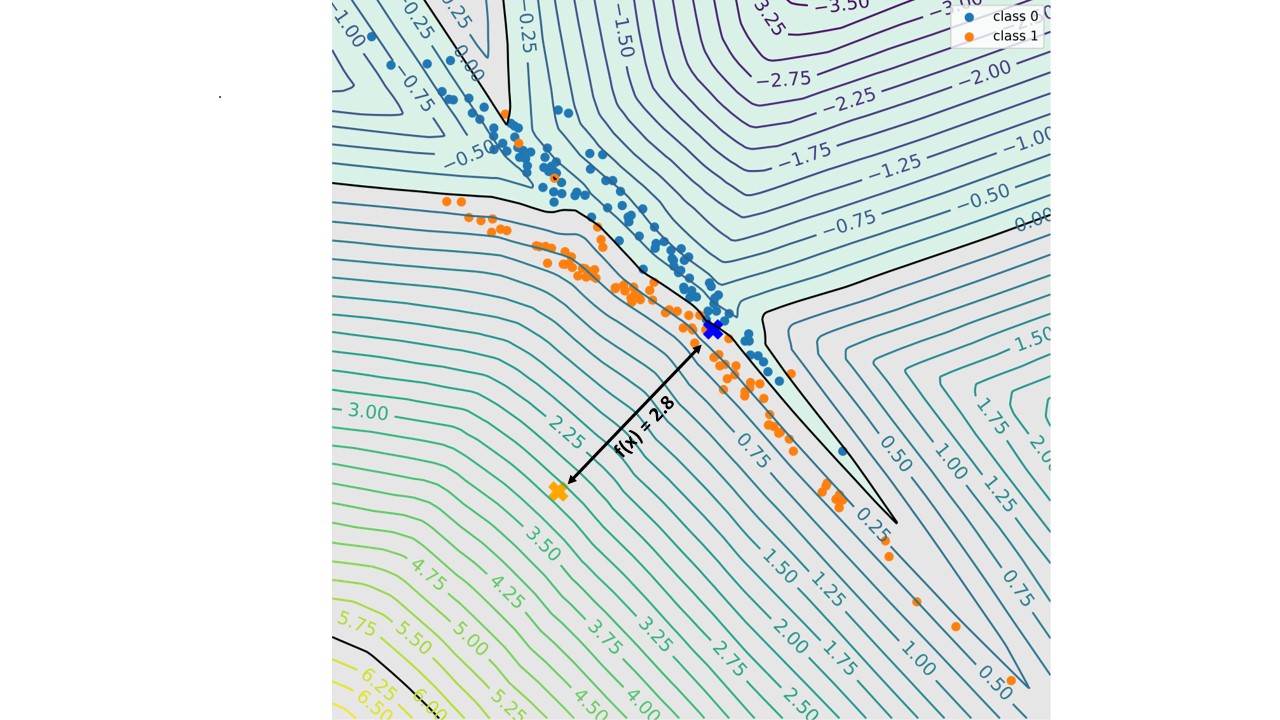}
    \caption{An example of a binary SDC. Observe that the countour lines are parallel-curves of the classification boundary (the black curve) and the output of the model in a $x$ (the orange cross) directly provides the distance from the closest point in the classification boundary (the blue cross).}
    \label{fig:toy-example}
\end{figure}

%\removeGiulio{
%Recently, \cite{brau} provided a practical fast approximation of the minimal distance perturbation with a theoretical estimation of the committed error, however, the approximation only works in a region close enough to the classification boundary.}
%-------------------------------------------
To achieve provable guarantees, other works focused on designing network models with bounded Lipschitz constant that, by construction, offers a lower bound of the MAP as the network output \cite{Tsuzuku_Sato_Sugiyama_2018}. These particular models can be obtained by composing orthogonal layers \cite{parseval, anil_conv, cayley,  optimal-transport, Singla_Feizi_2021} and norm-preserving activation functions, such as those presented by \cite{anil_fc, chernodub}. However, despite the satisfaction of the Lipschitz inequality, these models do not provide the exact boundary distance but only a lower bound that differs from the real distance.
%--------------------------------------------

\noindent
\textbf{This work} introduces a new family of classifiers, namely Signed Distance Classifiers (SDC), that \textit{straighten the Lipschitz lower bound by outputting the exact distance of an input $x$ from the classification  boundary}. SDC can then solve the MAP problem as a result of the network inference (see Figure \ref{fig:toy-example}).
From a theoretical point of view, we extend the characterization property of the signed distance functions to a multi-class classifier. From a practical perspective, we address such a theoretical model by presenting a new architecture, named Unitary-Gradient Neural Network (UGNN), \emph{having unitary gradient (under the Euclidean norm) in each differentiable point}. In summary, this work provides the following contributions:
\begin{itemize}
    \item It introduces a notable family of classifiers, named SDC, which provide as output the distance of an input $x$ from the classification boundary.
    \item It provides a novel network architecture, named UGNN, which, to best of our knowledge, represents the first practical approximation of an SDC.
    \item It shows that the $\abs$ function can replace other more expensive norm-preserving activation functions without introducing a significant accuracy loss. Furthermore, it proposes a new layer named \textit{Unitary Pair Difference}, which is a generalization of a fully-connected orthogonal layer.
    \item It assesses the performance, the advantages, and the limitations of the proposed architecture through a close comparison with the state-of-the-art models in the field of the Lipschitz-Bounded Neural Networks.
\end{itemize}
%\textbf{Structure:} after a discussion of the related work, we provide a definition and a characterization of the SDCs. %, first introducing the concept for a binary classifier and then extending i,t to a multi-class classifier. %, where a characterization property is discussed.
%Then, we propose the UGNN to practically address the required properties of an SDC. Finally, we discuss a set of experiments to evaluate UGNNs and discuss future works.
\section{Related Work}
The evaluation and the provable verification of the robustness of a classification model can be addressed by computing the MAP in a given point $x$ \cite{Carlini_Dill_2018}. Since that computation involves solving a minimum problem with non-linear constraints, the community focused on designing 
faster algorithms to provide an accurate estimation of the distance to the classification boundary \cite{rony_2019, rony_2020, pintor2021fast}. However, all these algorithms require multiple forward and backward steps, hence are not suited for an online application \cite{brau}.

On the other side, since the sensitiveness to input perturbations strictly depends on the Lipschitz constant of the model, knowing the local Lipschitz constant in a neighborhood of $x$ provides a lower bound of the MAP in $x$ \cite{Hein_Andriushchenko_2017}. In formulas, for a $L$-Lipschitz neural network $f$, a lower bound of MAP is deduced by considering$\frac{1}{L\sqrt{2}}(f_l(x) - f_s(x))$, where $l,s$ are the first and the second top-$2$ components, respectively. However, for common DDNs, obtaining a precise estimation of $L$ is still computationally expensive \cite{Weng_Daniel_2018a}, thus also this strategy is not suited for an online application.

For these reasons, recently, other works focused on developing neural networks with a bounded Lipschitz constant by design. %In particular on the $1$-Lipschitz neural networks that can be obtained by composing $1$-Lipschitz layers.
%Since the Lipschitz constant of  a fully-connected layer (w.r.t the euclidean distance) coincides with the spectral-norm of the matrix weight,
\cite{Miyato_Yoshida_2018} achieved $1$-Lipschitz fully connected layers by bounding the spectral-norm of the weight matrices to be $1$.
% considers fully-connected layers of the form $g(x) = \frac{W}{\|W\|} x +b$, where $\|W\|$ is the spectral norm of $W$.
Similarly, \cite{optimal-transport} considered neural networks $f$ in which each component $f_i$ is $1$-Lipschitz, thus, differently from the $1$-Lipschitz networks mentioned before, given a sample $x$, the lower bound of MAP is deduced by $\frac{1}{2}(f_l(x) - f_s(x))$.
%where $l,s$ are respectively the first and the second top-$2$ components.

Other authors leveraged \textit{orthogonal} weight matrices to pursue the same objective. For instance, \cite{orthDNN} showed that a ReLU Multi-Layer Perceptron merely composed by orthogonal layers is 1-Lipschitz. Indeed, an orthogonal matrix $W$ (i.e. such that $WW^T=I$ or $W^TW=I$) has a unitary spectral norm, $\|W\|=1$. Roughly speaking, orthogonal fully connected and convolutional layers can be obtained by \textit{Regularization} or \textit{Parameterization}.
The former methods include a regularization term in the training loss function to encourage the orthogonality of the layers, e.g \cite{parseval} use $\beta\|W^TW-Id\|^2$.
The latter methods, instead, consider a parameterization of the weight $W(\theta)$ depending on a unconstrained parameter $\theta$ so that, for each $\theta$, $W(\theta)$ is an orthogonal weight matrix \cite{anil_fc, cayley}.
For convolutional layers, a regularization strategy can be applied, since they can be written as matrix-vector product through a structured matrix \cite{toeplitz}.
%during the training stage in order to obtain a orthogonal-convolutional layer. 
However, recent parameterized strategies as BCOP \cite{anil_conv}, CayleyConv \cite{cayley}, and Skew Convolution \cite{Singla_Feizi_2021} come out as efficient and performant alternatives.   

\noindent
\textbf{This work} defines an SDC, as a function $f$ that provides the MAP by computing $f_l(x) - f_s(x)$, thus tightening the lower bounds provided by $L$-Lipschitz classifiers.
Furthermore, we present the UGNN, designed by properly leveraging the previous orthogonal parameterized strategies, as the first architecture that approximate a theoretical SDC.

%Furthermore, we will show that $f_l(x) - f_s(x)$ provides a more accurate (lower) estimation of the MAP in $x$ respect to the lower bounds of three of the state-of-the-art $1$-Lipschitz networks.

%Finally, \cite{mountains} theorized the existence of the \textit{Highest Mountain Functions} which feature similar behaviour of SDCs for the binary classification case.
%multi-classification case (see App.A, Def.4 of the paper). 
%However, they only focused on fully connected networks.
%\textbf{This work} instead, leverages orthogonal layers in order to build the proposed UGNN model $f$ that similar to the aforementioned networks features a bounded Lipschitz constant. However, 
%\cite{optimal-transport}

%From an operative point of view, the computational cost of finding MAP is transferred into the training stage.

\section{Signed Distance Classifier}

In this context, a classifier $\hat k:\XX\to\YY$ maps the input domain into a finite set of labels $\YY$. The concept of \textit{robustness} is formally stated in the following definition.

\begin{defn}[robustness]
  A classifier $\hat k$ is $\varepsilon$-robust in an input $x \in \R^n$ (or equivalently, a classification $\hat k (x)$ is $\varepsilon$-robust), if $\,\hat k(x+\delta) = \hat k(x)\,$ for any perturbation $\delta$  with $\|\delta\|<\varepsilon$, where $\|\cdot\|$ is the Euclidean norm.
\end{defn}

\subsection{Binary Classifiers}

Let $f:\R^n\to\R$ be a binary classifier that provides a classification of an input $x$ based on its sign, i.e., $\hat k(x) = \sgn(f(x))$, and let $\BB_f:=\left\{ x\in\R^n\,:\,f(x)=0 \right\}$ be the classification boundary of $f$. Given an input sample $x$, the \textit{MAP} problem for a binary classifier is defined as follows:
\begin{equation}
  \begin{aligned}
    d_f(x):= &\inf_{p\in\R^n} & \|p-x\|\\
           & \mbox{s.t.}    & f(p)=0,   
  \end{aligned}
  \label{eq:minimal-adversarial-problem}
\end{equation}
where $d_f$ represents the distance function from the boundary $\BB_f$.
The \textit{closest adversarial example} to $x$ is defined as the unique $x^*$ (if any) such that
$d_f(x) = \|x-x^*\|$ and $\sgn(f(x))\ne\sgn(f(x^*))$. Observe that 
%assumptions~\ref{ass:diff}~\ref{ass:regular}
Problem~\eqref{eq:minimal-adversarial-problem} is equivalent to the
definition of \textit{Minimal Adversarial Perturbation} in
\cite{deepfool}.

\smallskip
\noindent
\textbf{Certifiable robustness.} We refer to $\delta^*=x^*-x$ as the \emph{perturbation} that realizes the MAP. 
\add{By definition of $ d_f(x)$, for each perturbation $\delta$ with $\|\delta\|< d_f(x)$ it holds $\hat k (x+\delta)=\hat k(x)$; hence, $\hat k$ is certifiable $d_f(x)$-robust in $x$.}

\smallskip
A \textit{Signed Distance Function} $d^*_f$ is defined as follows:
\begin{equation}
  d^*_f(x)=
  \begin{cases}
    d_f(x) & x\in R_+\\
    -d_f(x) & x\not\in R_+.
  \end{cases}
  \label{eq:signed-distance}
\end{equation}
where $R_+=\{f>0\}$.
Following this definition, a signed distance function $d^*_f$ satisfies intriguing properties that make it highly interesting for \textit{robustness evaluation, verification, and certifiable prediction}. 
In particular, $d^*_f$ provides the same classification of $f$, since $\sgn (d_f^*(x)) = \sgn (f(x))$ for each $x\in\R^n$. Furthermore, the gradient $\nabla d^*_f(x)$ coincides with the direction of the shortest path to reach the closest adversarial example to $x$ \cite[Thm. 4.8]{federer}.

%Finally, $d_f^*(x)$ represents an \textit{online certification of the classification $\hat %5k(x)$, indeed
%Moreover, given an $\varepsilon >0$, the classifier $\hat k_\varepsilon$ defined by
%\begin{equation}
%    \forall x\in\R^n,\quad
%    \hat k_\varepsilon = 
%    \begin{cases}
%        \hat k(x) & d_f(x) >0 \\
%        0 &\mbox{otherwise}
%    \end{cases}
%\end{equation}
%\todo{Add here a concise definition of robust classifier and how to obtain it from the solution of the MAP problem and hence from $d^*_f(x)$.}

\begin{oss}
  \label{oss:sign-dist-grad}
  Let $x\in\R^n$, if there exists a unique $x^*\in\BB_f$ such that $d_f(x) = \|x-x^*\|$, then $d_f^*$ is differentiable in $x$ such that
  \begin{equation}
    \nabla d_f(x) = \frac{x-x^*}{\|x-x^*\|},
  \end{equation}
    and hence has a gradient with unitary Euclidean norm, i.e., $\|\nabla d_f^*(x)\|=1$, referred to as \emph{unitary gradient} (UG) for short in the following.
  Furthermore, $d^*_f$ is such that:
  \begin{enumerate}
    %\[
    %  \CC(f):=\left\{ g:\R^n\to\R\,:\, \sgn(g)\equiv\sgn(f) \right\}
    %\]
    %provide the same classification of $f$, and, by construction,
    %$d^*\in\CC(f)$. Moreover, two functions $f,\,g$ provide the same
    %classification if and only if $d^*_f\equiv$<++>
    \item \label{enum:safe} It provides a trivial way to
      certify the robustness of \add{$\hat k$ in} $x$, since, by definition, $|d^*_f(x)|$ represents the  MAP. 
    \item \label{enum:explicit} It explicitly provides the closest adversarial example to $x$, which can be computed $x^* = x - d^*_f(x) \nabla d^*_f(x)$.
  \end{enumerate}
  \begin{proof}
    Refer to \cite[Thm. 4.8]{federer}
  \end{proof}
\end{oss}
Inspired by these intriguing properties, this work aims at
\textit{investigating classifiers whose output provides the 
  distance (with sign) from their own classification boundary.} 

\subsection{A Characterization Property}

%This section provides a characterization of the signed
%distance functions $d_f^*$. 
A trivial example of a binary classifier $f$ that coincides with a signed distance function is given by any \textit{affine function} with a unitary weight. Indeed, if $f(x) = w^Tx+b$, where $\|w\|=1$, then the MAP relative to $f$
has the explicit unique solution of the form $x^* = x -
\frac{f(x)}{\|w\|^2}w$, as already pointed out in \cite{deepfool}, from which $d_f(x) = |f(x)|$.

As shown in Observation~\ref{oss:sign-dist-grad}, a signed distance function has a unitary gradient. Under certain hypotheses, the opposite implication holds: a function $f$ with a unitary gradient coincides with a signed distance function from $\BB_f$. This result is formalized in the following theorem.
\begin{thm}
  \label{thm:sdf-characterization}
  Let $\UU\subseteq\R^n$ be an open set, 
  and let $f:\R^n\to\R$ be a function, smooth in $\UU$, such that $\BB_f\subseteq\UU$. If $f$ has a unitary gradient in $\UU$, then 
  there exists an open set $\Omega_f\subseteq\UU$ such that $f$ coincides in $\Omega_f$ with the signed distance function from $\BB_f$. 
  Formally, 
  \begin{equation}
      \|\nabla f_{\restriction\UU} \| \equiv 1 \quad \Rightarrow \quad\exists\Omega_f\subseteq\UU,\quad  f_{\restriction \Omega_f} \equiv d^*_{f\restriction \Omega_f}.
      \label{eq:unitary-constraint}
  \end{equation}
    \begin{proof}
    The proof is built upon \cite[Prop.2.1]{sakai}.
    Any trajectory $\gamma:[0,1]\to\UU$ that solves the dynamical system $\dot \gamma(t) = \nabla f(\gamma(t)))$ coincides with the shortest path between the point $\gamma(0)$ and the hyper-surface
    $f^{-1}(\gamma(1))$. Details are reported in the Appendix. 
  \end{proof}
\end{thm}
 \iffalse
 %%%%% OLD VERSION %%%%
\begin{thm}
  \label{thm:sdf-characterization}
  Let $\UU\subseteq\R^n$ be an open set, 
  and let $f:\R^n\to\R$ be a function, smooth in $\UU$, such that $\BB_f\subseteq\UU$. Then, there exists an open set $\Omega_f\subseteq\UU$, such that $f$ has a unitary gradient in $\Omega_f$ 
  if and only if $f$ coincides, in $\Omega_f$, with the signed distance function with respect to $\BB_f$. 
  Formally,
  \begin{equation}
      \|\nabla f_{\restriction\Omega_f} \| \equiv 1 \quad \iff \quad f_{\restriction \Omega_f} \equiv d^*_{f\restriction \Omega_f}.
  \end{equation}
  \begin{proof}
    The proof is built upon \cite[Prop.2.1]{sakai}.
    Any trajectory $\gamma:[0,1]\to\UU$ that solves the dynamical system $\dot \gamma(t) = \nabla f(\gamma(t)))$ coincides with the shortest path between the point $\gamma(0)$ and the hyper-surface
    $f^{-1}(\gamma(1))$. Details are reported in the Appendix. 
  \end{proof}
\end{thm}
\fi
It is worth noting that, as pointed out in \cite[Prop.2.1]{sakai},
this characterization holds for particular geometrical spaces, i.e., 
\textit{Complete Riemannian Manifolds}. Unfortunately, as shown by the author, the only smooth functions with unitary gradient in a Complete Remannian Manifold with 
non-negative Ricci Curvature (e.g., $\R^n$) are the affine functions \cite[Theorem A]{sakai}. 
However, an open set $\UU\subset\R^n$ is a Remannian Manifold that 
does not satisfy the completeness property. 
Hence, the existence of a non-affine signed distance function 
is not in contradiction with \cite[Theorem A]{sakai}. A trivial example is given by the binary classifier $f(x) = \|x\|-1$ defined in $\UU=\R^n\setminus\{0\}$. 
Further details are provided in the Appendix.
%\todo
%Observe in conclusion that, from the point of view of the Variational Calculus  a signed distance function correspond to the solution of an Eikonal equation. Which is a widely known differential equation involved in the analysis of the propagation speed of optic and acoustic waves in presence of different physical mediums.

\subsection{Extension to Multi-Class Classifiers}
%The characterization above only considers binary classifiers.
%However, b
By following the \textit{one-to-rest} strategy \cite{smola}, the results above can be extended to multi-class classifiers. Let $f:\R^n\to\R^C$ be a smooth function by which the predicted class of a sample
$x\in\R^n$ is given by $\hat k(x) = \argmax_i f_i(x)$, where $\hat k(x)=0$ if there is no unique maximum component. Observe that, according
to \cite{biggio, szegedy, deepfool}, the MAP problem for a multi-class classifier
%, 
%i.e., the problem of of finding the %closest adversarial example to $x$,
can be stated as follows:
\begin{equation}
  \begin{aligned}
    d_f(x):= &\inf_{p\in\R^n} & \|p-x\|\\
    & \mbox{s.t.} & \hat k(p) \ne \hat k(x).
  \end{aligned}
  \tag{MAP}
  \label{eq:map}
\end{equation}
Here, we extend the definition of signed distance function $d^*_{f}$ to a multi-class \textit{Signed Distance Classifier} $f$ as follows.
%The motivation behind the following definition will be clear in the
\begin{defn}[Signed Distance Classifier]
  A function $f:\R^n\to\R^C$ is a \textit{Signed Distance Classifier} 
  if, for each pair $i,j$, with $i\ne j$, the
  difference $(f_i-f_j)$ corresponds to the signed distance
  from the one-to-one classification boundary $\BB_{ij}:=\{x\in\R^n\,:\,f_i-f_j=0\}$.
\end{defn}

The following observation shows that an SDC satisfies similar properties 
%to~\ref{enum:safe} and~\ref{enum:explicit} 
of Observation~\ref{oss:sign-dist-grad} for binary classifiers. 
\begin{oss}
  \label{oss:multi-motivation}
  Let $f:\R^n\to\R^C$ be a signed distance function and let $x\in\R^n$ be a sample classified as
  $l = \hat k(x)$. Let $s:=\argmax_{j\ne l} f_j(x)$ be the second-highest component of $f(x)$. Hence, the classifier $f$:
  \begin{enumerate}
  \item Provides a fast way to certificate the
    robustness of \add{$\hat k$ in}  $x$. In fact, $f_l(x) - f_s(x) = d_f(x)$, where $d_f(x)$ is the MAP.
  \item Provides the closest adversarial example to $x$, i.e., 
    \[
      x^* = x - (f_l(x)-f_s(x)) \nabla (f_l-f_s)(x),
    \]
    where $x^*$ is the unique solution of Problem~\ref{eq:map} in $x$.
 \end{enumerate}
 \begin{proof}
   The detailed steps are in the Appendix.
 \end{proof}
\end{oss}

Similarly to the binary case, an SDC is characterized by having a \emph{unitary gradient} for each pair-wise difference of the output components. In details, by directly applying Theorem~\ref{thm:sdf-characterization}, a smooth classifier $f$ is a signed distance classifier (in some open set) if and only if $\|\nabla (f_i-f_j)\|\equiv 1, \forall i\ne j$.
%% already defined above!!
%In the following, this is referred to as the unitary gradient (UG) property.}
%, namely \textit{Unitary Gradient} property.

\section{Unitary-Gradient Neural Networks}

In the previous section, we showed that if a smooth classifier $f$ satisfies the unitary gradient property in some open set $\UU\supseteq\BB_f$, then it admits an open set $\Omega_f\supseteq\BB_f$ in which $f$ fcoincides with the signed distance function with respect to the boundary $\BB_f$. Furthermore, affine functions represents all and the only smooth SDCs in the whole $\R^n$.
%---------------------------------------------------
%GB: Ho cambiato "SDCs smooth" in "smooth SDCs" perché altrimenti la frase non aveva senso. A meno che non manca qualcos'altro.
%--------------------------------------------------- checked

Supported by these results, any DNN that globally satisfies the UG property would coincide with a trivial linear model, which hardly provides good classification performance for complex tasks. To approximate a non-trivial SDC with a well-performing DNN $f_\theta$, we impose the UG property almost-everywhere.%, by relaxing the constraint on regularity.
%Namely, let $\HH:=\{f_\theta\,:\, \jac(f_\theta)\jac(f_\theta)^T = \frac{1}{2}Id\}$.
This section shows the proper requirements on $f_\theta$ to satisfy the hypothesis of Theorem~\ref{thm:sdf-characterization},
providing layer-wise sufficient conditions that ensure the UG property.
To this end, we focus our analysis on the family $\FF$ of feed-forward DNNs $f:\R^n\to\R^C$ of the form $f=g\circ h^{(L)}\circ\cdots \circ h^{(1)}$, where $g$ is the output-layer and each $h^{(i)}$ is any canonical elementary layer (e.g., Fully Connected, Convolutional, etc.) or an activation function.

%---------------------------------------------------
%GB: Più che una Observation, questo non sarebbe meglio un Lemma?
%---------------------------------------------------
\begin{oss}[Layer-wise sufficient conditions]
    \label{oss:layer-suff-conditions}
  Let $f$ be a DNN in $\FF$. For each $i$, let $J^{(i)}(x)$ be the Jacobian of $h^{(i)}$ evaluated in $y=h^{(i-1)}\circ\dots\circ h^{(1)}(x)$. For each $j=1,\ldots,C$, 
  let $W_j(x)$ be the Jacobian of $g_j$ evaluated in $y=h^{(L)}\circ\dots\circ h^{(1)}(x)$. Hence, if    \begin{equation}
    J^{(i)}J^{(i)T} \equiv I, \quad\forall i=1,\dots,L 
    \tag{GNP}
    \label{eq:orth-constr}
  \end{equation}
  and 
  \begin{equation}
      (W_h - W_k)(W_h - W_k)^T\equiv 1,\quad\forall h\ne k,
    \tag{UPD}
  \end{equation}
  then, for all $h\ne k$, $f_h - f_k$ satisfies the UG property.% of the Theorem~\ref{thm:sdf-characterization}.
  \begin{proof}
    For a feed-forward neural network, the Jacobian
    matrix of each component $f_j$ can be decomposed as
    \begin{equation}
      \jac(f_j) = W_j\prod_{i=1}^L J^{(i)}=W_j J^{(L)}\cdots  J^{(1)}.
      \label{jac-feedforward}
    \end{equation}
    Hence, the thesis follows by the associative property and by observing 
    that $(AB)^T=B^TA^T$ for any two matrices.
  \end{proof}
\end{oss}
Observe that Condition~\ref{eq:orth-constr}, namely \textit{Gradient Norm Preserving}, requires any layer to have an output dimension no higher than the input dimension. Indeed, a rectangular matrix $J\in\R^{M \times N}$ can be orthogonal by row, i.e, $JJ^T=I$, only if $M\le N$. Condition GNP is also addressed in \cite{anil_conv, cayley} to build Lipschitz-Bounded Neural Networks. However, for their purposes, the authors also consider DNNs that satisfy a weaker condition named \textit{Contraction Property} (see \cite{cayley}), which includes the $M\ge N$ case.
\subsection{Gradient Norm Preserving Layers}
We now provide an overview of the most common layers that can satisfy the GNP property. For a shorter notation, let $h$ be a generic internal layer.
\paragraph{Activation Function}
Activation functions $h:\R^n\to\R^n$ can be grouped in two main categories: \textit{component-wise} and \textit{tensor-wise} activation functions. 
Common component-wise activation functions as ReLU, tanh, and sigmoid do not satisfy the GNP property \cite{chernodub}. Moreover, since any component-wise function $h$ that satisfies the \ref{eq:orth-constr} property is piece-wise linear with slopes $\pm 1$ (see the appendix for further details), $\abs$ is GNP.% since $\jac(h)(x) = \diag(\alpha_1,\cdots,\alpha_n)$ where $\alpha_i\in\{-1,1\}$.

Tensor-wise activation functions have recently gained popularity
thanks to \cite{chernodub, anil_fc, singla2021improved}, who introduced OPLU, GroupSort, HouseHolder activation functions, respectively, %The OPLU activation is a particular case of the GroupSort. 
which are specifically designed to satisfy the \ref{eq:orth-constr} property. 
An overview of these activation functions is left in the appendix. In this work, we compare the $\abs$ with the OPLU and the GroupSort with a group size of $2$, a.k.a MaxMin. 
%\todo{AB: arrived here}
\paragraph{Fully Connected  and Convolutional Layers}
A fully connected layer of the form $h(x)=Wx+b$ has a constant Jacobian matrix $\jac(h)(x)=W$. This implies that $h$ is GNP if and only if $W$ is an orthogonal-by-row matrix. 
Similarly, for a convolutional layer with kernel $\KK$ of shape $M\times C \times k\times k$, the GNP property can be satisfied only if $M\le C$, i.e., the layer does not increase the number of channels of the input tensor~\cite{orthDNN}.
%TODO
As done in \cite{anil_fc}, in our model we consider the Bj\"{o}rck parameterization strategy to guarantee the orthogonality of the fully connected layers and the CayleyConv strategy presented in \cite{cayley} for the convolutional layers.

\paragraph{Pooling, Normalization and Residual Layers}
%\todo{Appendix}
Max-pooling two-dimensional layers with kernel $\bm k=(k_1,k_2)\in\N^2$, stride $\bm s=\bm k$, and without padding, satisfy the GNP property if applied to a tensor whose spatial dimensions $H, W$ are multiples of $k_1$ and $k_2$, respectively. This can be proved by observing that the Jacobian matrix corresponds to an othogonal projection matrix \cite{orthDNN}. 

Batch-normalization layers with a non-unitary variance do not satisfies the GNP property \cite{orthDNN}. For residual blocks, it is still not clear whether they can or cannot satisfy the GNP property. Indeed, a residual layer of the form $h(x) = x+\tilde h(x)$ is GNP if and only if $\jac(\tilde h)\jac(\tilde h)^T +\jac(\tilde h) + \jac(\tilde h)^T\equiv 0$. For such reasons, the last two mentioned layers are not considered in our model. 
%Indeed, if $h$ is such Average Pooling  is applied to an input tensor of shape $c\times n\times m$, then it can be written in form of a convolution $K\in\R^{c\times c\times k\times k}$ such that $K[i,i]\equiv\frac{1}{k^2}$ and such that $K[i,j]\equiv 0$ for $i\ne j$. Hence observe that $h(x)=K \star x$. By leveraging the structure of the corresponding matrix $\KK$, we can prove that $\KK\KK^T = \frac{1}{k^2}Id\in\R^{cnm\times cnm}$. 
%Similarly, a maxpool layer $h$ applied to an input tensor $x\in\R^{c\times n\times m}$ is such that the Jacobian tensor $P$ of $h$ has a block-diagonal $P=\diag\left(P_1,\cdots,P_c \right)$ where each diagonal-block has shape $n'm'\times nm$. Observe that each $P_i$ is a projection matrix, i.e., it has one single $1$ in each row and at most a $1$ entry in each column. Based on this observation, we can deduce that $P_iP_i^T=Id$ from which $PP^T=Id$.

\subsection{Unitary Pair Difference Layers}

%While the previous section analyzed the Gradient Norm Preserving property, 
This section focuses on the second condition stated in Observation~\ref{oss:layer-suff-conditions}: the \textit{Unitary Pair Difference} (UPD).

Since most neural classifiers include a last fully-connected layer, we restrict our analysis to this case. Let $g(x)=Wx+b$ be the last layer, since $\jac(g) \equiv W$, then the UPD property requires that for each two rows $W_h,\,W_k$ the difference $W_h-W_k$ has unitary norm. Let us say that a matrix $W$ satisfies the UPD property if the function $x\mapsto Wx$ is UPD. 

\noindent
\textbf{Bounded UPD layer.} An UPD matrix from any orthogonal-by-row matrix as stated by the next observation.
\begin{oss}
    Let $Q\in\R^{m\times C}$ such that $QQ^T=I$. Then,  $W=\frac{1}{\sqrt{2}}Q$ satisfies the UPD property, indeed
    \begin{equation}
      \|W_h- W_k\|^2=\underbrace{\|W_h \|^2}_{1/2}+\underbrace{\|W_k\|^2}_{1/2}-2 \underbrace{W_h^T W_k}_{0} = 1.
    \end{equation}
\end{oss}

An UPD layer with matrix $W$ as above is said to be \emph{bounded}, as each row of $W$ is bounded to have norm $1/\sqrt{2}$. 

As pointed out in \cite{singla2021improved}, this constraint makes it harder to train the model when the output dimension $C$ is large (i.e., there are many classes).

\noindent
\textbf{Unbounded UPD layer.}
To avoid this issue, we considered an UPD layer with parametric weight matrix $W(U)$. % of the same dimensions of $W$. 
Matrix $W(U)$ is obtained by iteratively applying the L-BFGS, an optimization algorithm for unconstrained minimum problems \cite{liu1989limited}, to the loss 
\begin{equation}\label{eqn:psi}
 \Psi(U) = \sum_{h< k} (\|U_h-U_k\|^2 - 1)^2.
\end{equation}
More specifically, if $\texttt{psi}$ is the routine that computes such a loss function and $\texttt{L-BFGS}$ is the routine that performs one step of the L-BFGS optimization algorithm,
then the weight matrix is obtained as $W = \texttt{UPD}(U)$, where \texttt{UPD} is the following procedure:
\begin{lstlisting}[language=Python, caption={Psudo code that parameterizes an UPD matrix through a parameter $U$. The resulting $W$ is obtained by performing few steps of the L-BFGS algorithm to find a minimum of $\Psi$ with starting point $U$.}, captionpos=b]
def UPD(U: Tensor):
    # Returns an UPD matrix
    W = U
    for in range(max_iter):
        W = L-BFGS(psi(U), W)
    return W
\end{lstlisting}

Note that the UPD layer $g(x) = W(U)\, x + b$ depends on the weights $U,b$ and it is fully differentiable in $U$. This implies that such a procedure, like parameterization strategies for orthogonal layers, can be applied during training. 
Finally, note that the algorithm complexity strongly depends on the computational cost of the objective $\Psi(U)$. Our implementation exploits parallelism by implementing $\Psi(U)$ by means of a matrix product of the form $A^{(C)}U$, where $A^{(C)}$ is designed to compute all the pair differences between rows required by Eq.~\eqref{eqn:psi} (see Appendix).
\iffalse
\subsection{Training and Loss Function}
From a ML point of view, we can leverage the aforementioned sufficient
conditions to practically craft a Signed Distance Neural Network. 
Let $\XX=(x,y)\in\R^n\times\{1,\cdots,C\}$ be a distribution of data, let $\LL$ a loss
function, and let $f_\theta$ a family of neural network parameterized by
$\theta$. Searching for a $\theta^*$ such that $f=f_{\theta^*}$ is a signed
distance neural network which minimize the empirical loss over $\XX$ involves
the solution of the following minimum problem 
\begin{equation}
  \begin{aligned}
    &\min_{\theta} &\E_\XX\left[\LL(f_\theta(x),y)\right]\\
    &\mbox{s.t.}   &\jac(f_\theta)\jac(f_\theta)^T \equiv_{m} \frac{1}{2}Id
  \end{aligned}
  \tag{SDN}
  \label{eq:sdn-minmum}
\end{equation}
where $\equiv_{m}$ means that the equivalence holds for all the points
except a set of zero measure.
\label{thm:equivalence}
\fi

\subsection{Unitary-Gradient Neural Network Architecture}
\begin{figure}[ht]
    \centering
    \includegraphics[width=\columnwidth, keepaspectratio]{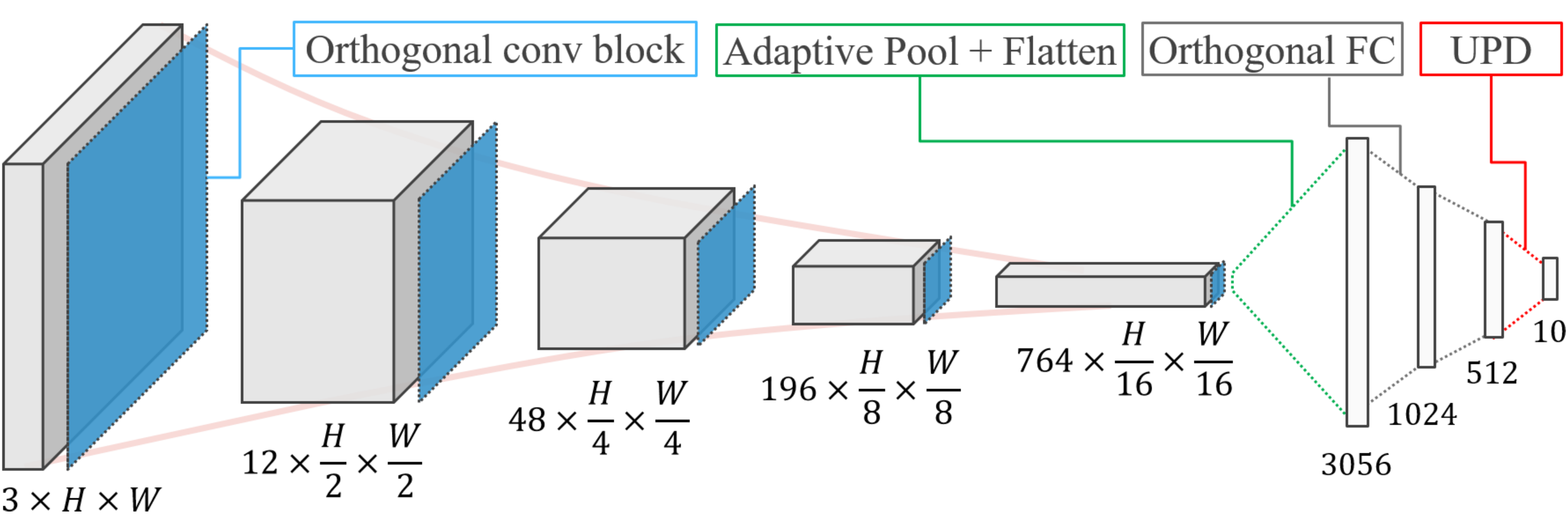}
    %\caption{TODO-G: I suggest pointing out an examples of the decreasing number of neurons -> "For instance, looking the EUNet architecture, given an image 3x32x32, the number of neurons decreases at each block as follow: 3x32x32, 12x16x16, 48x8x8,..., which allows respecting the architectural assumption of a sgd Net".}
    \caption{Tested UGNN architecture: 5 GNP conv-blocks, 2 FC GNP layers and 1 UPD layer.}
    %Since the convolutional layers have to preserve the input channels, the pixel-unshuffle layers at the end of each block are used to increase the feature dimensions.
    \label{fig:architecture}
\end{figure}

This section describes how to practically combine GNP and UPD layers to obtain a neural network $f_\theta$ such that all  pair-wise differences of its output vector \add{have} unitary gradient. 
The main difficulty in crafting such a network is due to the GNP property, which implies a decreasing dimension in both dense and convolutional layers.
%In details, the kernel of a generic internal convolution, of shape $M\times N\times k_1\times k_2$, that satisfies GNP property necessarily is bounded to have $M\le N$. 
Indeed, most DNNs for image classification process a $3$-channel image by gradually increasing the channel dimension of convolutional layers.

To overcome this issue, we leverage a 2-dimensional \textit{PixelUnshuffle} layer \cite{shi2016real}, which inputs a tensor of shape $C \times rH\times rW$ and outputs a tensor of shape $r^2C\times H\times W$. The output is obtained by only rearranging input components. As such, this layer satisfies the GNP property (proof available in Appendix). The main advantage of using a PixelUnshiffle layer is that it allows increasing the number of channels of hidden layers even in convolutional GNP networks.

\begin{table}[h!]
    \centering
    \begin{tabular}{lc}
    \textbf{Layers} & \textbf{Output Shape} \\
    \midrule
        \texttt{OrthConv}$(3\cdot4^i, 3\cdot4^i, 3)$ & $3\cdot 4^i\times \frac{H}{2^i} \times \frac{H}{2^i}$\\
        \texttt{GNP Activation}  & " \\
        \texttt{OrthConv}$(3\cdot4^i, 3\cdot4^i, 3)$ & "\\
        \texttt{GNP Activation}  & "\\
        \texttt{PixelUnshuffle}$(2)$ & $3\cdot 4^{i+1}\times \frac{H}{2^{i+1}} \times \frac{H}{2^{i+1}}$\\
    \end{tabular}
    \caption{An example of the (i+1)th internal GNP conv-block. Observe that the number of channels increases with a geometric progression of common ratio $4$ and each spatial dimension decreases with a ratio of $2$.}
    \label{tab:conv-block}
\end{table}
It is worth pointing out that \cite{anil_conv, cayley} also leveraged such a permutation layer, but only to emulate a convolution with stride $2$.
That said, the UGNN proposed in this work, shown in Fig.~\ref{fig:architecture}, consists of five GNP blocks, two fully connected GNP layers, a last UPD layer (bounded or unbounded), and GNP activation functions. Each GNP block consists of two GNP convolutional layers and one last PixelUnshuffle layer with scaling factor $2$; a GNP activation function is applied after each convolution (see Tab. \ref{tab:conv-block}).  Each convolutional layer has a circular padding to preserve the spatial dimension of the input. Furthermore, before the flattening stage, a max-pool layer with window size and stride $H/2^5$ is applied to process input of different spatial dimension $H=m\cdot 2^5$, for any $m\in\N$.

Note that, to the best of our records, this is the first instance of a convolutional DNN that aims at pratically implementing an SDC and that provably satisfies $\|\nabla (f_i-f_j)\|\equiv 1$ almost everywhere. \cite{mountains} only focused on fully-connected networks, while \cite{optimal-transport} only approximated an optimal $f^*$ such that $\|\nabla f^*_i\|\equiv 1$.

In conclusion, observe that, by design, each pair-difference $f_i-f_j$ of an UGNN satisfies the $1$-Lipschitz property, hence the margin $\MM_f(x) = f_l(x) - \max_{j\ne l}f_j(x)$ is a lower bound of the MAP in $x$.
\begin{oss}[Certifiable Robustness]
\label{oss:robustness}
%Let $f$ be an UGNN and $\hat k(x)=l$, for each perturbation $\delta$ such that $\|\delta\|\le\MM_f(x)$ the perturbed input $\hat k(x+\delta) = \hat k(x)$. In other terms, $\MM_f(x)$ is directly a lower bound of the MAP in $x$. 
\add{If $f$ is a UGNN, then $\hat k(x) = \argmax_i f_i(x)$ is $\MM_f(x)-$robust in $x$.}
%In other words, $\MM_f(x)$ is directly a lower bound of the MAP in $x$.}
\begin{proof}
The proof in available in Appendix.
\end{proof}
\end{oss}
%Finally, under such guarantees, given any $\varepsilon>0$, the classifier $\hat k_\varepsilon$, obtained by rejecting the samples with a $\MM_f$ lower than $\varepsilon$, is provably robust.
\iffalse
\subsection{Provable Robustness of an UGNN}
%As stated in \cite{anil_conv}, if is $f$ $1$-Lipschitz, and $x$ such that $\hat k(x)=l$, then $x$ is provably robust against perturbation smaller than $\MM_f(x) /\sqrt{2}$, where $\MM_f(x) = f_l(x) - \max_{j\ne l}f_j(x)$, that hence is a lower bound of the MAP in $x$. 
By definition, an UGNN $f$ is such that for each $i,j$ the difference $f_i - f_j$ is stronger than the lipschitz property of $f_i-f_j$.

%implies that each component difference $f_i-f_j$ is $1$-Lipschitz and so the following result holds implies a different lower-bound in the term of $\MM_f$.
\begin{oss}
\label{oss:robustness}
Let $f$ an UGNN and $\hat k(x)=l$, for each perturbation $\delta$ such that $\|\delta\|\le\MM_f$ the perturbed input $\hat k(x+\delta) = \hat k(x)$. In other terms, $\MM_f(x)$ is directly a lower bound of the MAP in $x$. 
\begin{proof}
The details are left in the appendix.
\end{proof}
\end{oss}
Finally, under such guarantees, given any $\varepsilon>0$, the classifier $\hat k_\varepsilon$, obtained by rejecting the samples with a $\MM_f$ lower than $\varepsilon$, is provably robust.
%: If $\|\delta\|\le\frac{\MM_f(x)}{\sqrt{2}}$, $\hat k(x) = \hat k(x+\delta)$ 
\fi
\section{Experimental Results}
Experiments were conducted to evaluate the classification accuracy of a UGNN and its capability of implementing an SDC. As done by related works, the experiments targeted the CIFAR10 datasets. We compared UGNN with the following $1$-Lipschitz models: \textit{LargeConvNet} \cite{anil_conv}, \textit{ResNet9} \cite{cayley}, and \textit{LipConvNet5} \cite{Singla_Feizi_2021}. For all the combinations of GNP activations, UPD layers, preprocessing, and input size, our model was trained for 300 epochs, using the Adam optimizer \cite{adam}, with learning rate decreased by $0.5$ after $100$ and $200$ epochs, and a batch of $1024$ samples, containing randomly cropped and randomly horizontally flipped images.
The other models were trained by following the original papers, leveraging a multi-margin loss function with a margin $m=\varepsilon\sqrt{2}$, with $\varepsilon=0.5$.
For a fair comparison, UGNN was trained with a margin $m=\varepsilon$, being the lower bound $\MM_f$ of the MAP for UGNN different from the $\MM_f/\sqrt{2}$ of the other DNNs, as discussed in Observation~\ref{oss:robustness}.
For the experiments, we used $4$ Nvidia Tesla-V100 with cuda 10.1 and PyTorch 1.8 \cite{pytorch}.
\subsection{Accuracy Analysis}
Table~\ref{tab:accuracy} summarizes the accuracy on the testset, where UGNN was tested with the ({Abs, MaxMin, OPLU}) activation, and the last UPD layers (bounded and unbounded).
\begin{table}[h]
    \centering
    \begin{tabular}{lcr}
\toprule
{} & \multicolumn{2}{c}{\bf Accuracy [\%]} \\
\textbf{Models} &   \textbf{Std.Norm} & \textbf{Raw} \\
\midrule
LargeConvNet    &  $\textbf{79.0}\pm0.26$ &  $\textbf{72.2}\pm0.11$ \\
LargeConvNet+Abs &  $77.8\pm0.33$  & $71.8\pm0.25$\\
LipConvNet5    &  $78.0\pm0.26$ &  $68.8\pm0.35$ \\
LipConvNet5+Abs &  $76.1\pm0.31$ & $65.5\pm0.68$ \\
ResNet9         &  $78.7\pm0.22$ &  $66.4\pm0.17$ \\
ResNet9+Abs     &  $78.1\pm0.34$ & $65.6\pm0.22$  \\      
\midrule
UGNN+Abs+updB    &  $71.9\pm0.29$ &  $69.2\pm0.31$ \\
UGNN+Abs+updU     &  $72.1\pm0.54$ &  $68.9\pm0.81$ \\
UGNN+MaxMin+updB &  ${72.6}\pm0.79$ &  $70.4\pm0.52$ \\
UGNN+MaxMin+updU  &  $\textbf{72.7}\pm0.38$ &  $70.4\pm0.86$ \\
UGNN+OPLU+updB   &  $71.9\pm0.09$ &  ${70.5}\pm0.39$ \\
UGNN+OPLU+updU    &  $72.0\pm0.70$ &  $\textbf{70.6}\pm0.45$ \\
\bottomrule
\end{tabular}
    \caption{Accuracy comparison between the 1-Lipschitz models and the UGNNs.}
    \label{tab:accuracy}
\end{table}
The other models were tested with the original configuration and with the $\abs$ activation. Experiments were performed \add{with and without} standard normalization (Std.Norm) of the input, and each configuration was trained four times with randomly initialized weights to obtain statically sound results.
In summary, the take-away messages of the Tab.~\ref{tab:accuracy} are:
\begin{enumerate*}[label=(\roman*)]
    \item The unbounded UPD layer (named updU) increased the performance with respect to the bounded one (named updB) in almost all cases.
    \item Std.Norm pre-processing significantly increased the performance. We believe this is due to the GNP property of the layers, which cannot learn a channel re-scaling different from $\pm 1$.
    \item The use of $\abs$ activations in the $1$-Lipschitz models does not cause a significant performance loss with respect to the other GNP activations (that requires a more expensive sorting).
    %\item Despite the strict constraints of the UGNN architecture, the performance can still be compared with the others $1$-Lipschitz models.
    \item Despite the strict constraints of the UGNN architecture, it achieves comparable performance in the raw case, while there is a clear gap of accuracy for the Std.Norm case.
\end{enumerate*}

To improve the UGNN accuracy, we investigated for intrinsic learning characteristics of its architecture. In particular, we noted that a strong limitation of the model is in the last two GNP blocks (see Fig.~\ref{fig:architecture}), which process tensor with a high number of channels (thus higher learning capabilities) but with compressed spatial dimensions ($H/8$ and $H/16$). 
Hence, for small input images (e.g., $32\times 32$), such layers cannot exploit the spatial capability of convolutions. Table~\ref{tab:accuracy-input-size} reports a performance evaluation of the UGNN (with MaxMin activation) for larger input sizes. Note that, differently from the UGNN, common DNNs do not benefit of an up-scaling image transformation, since it is possible to apply any number of channels on the first convolutions layers. Moreover, the compared models do not have adaptive layers, hence, do not handle different input sizes. This observation allows the UGNN to outperform the other models for the raw case and reach similar accuracy for the Std.Norm case.

\begin{table}[h]
    \centering
    \begin{tabular}{lllc}
\toprule
   & {} & \multicolumn{2}{c}{\bf Accuracy [\%]} \\
\textbf{Input Size} & \textbf{Last Layer} &           {\bf Std.Norm} &            {\bf Raw} \\
\midrule
%32 & orth &  $72.6\pm0.79$ &  $70.4\pm0.52$ \\
%   & upd &  $72.7\pm0.38$ &  $70.4\pm0.86$ \\
64 & updB &  $72.1\pm0.27$ &  $72.4\pm0.42$ \\
   & updU &  $72.6\pm0.69$ &  $72.8\pm0.61$ \\
128 &updB &  $74.5\pm0.56$ &  $75.9\pm0.07$ \\
   & updU &  $74.9\pm0.45$ &  $76.2\pm0.30$ \\
256 & updB & $76.5\pm0.35$ &  $78.4\pm0.29$ \\
   & updU &  $\textbf{76.8}\pm0.29$ & $\textbf{78.5}\pm0.22$ \\

\bottomrule
\end{tabular}

    \caption{Accuracy comparison of the UGNN models with different,
    pre-processed input sizes and output layers.
    %datasets, different input sizes and different output layers.
    }
    \label{tab:accuracy-input-size}
\end{table}

%Note that we did not compare on purpose UGNN with the $1$-Lipschitz models, since the aim of this test is to show that \textit{the accuracy of the UGNN increases with the size of the input images}.
%We believe this is due to the last two GNP blocks in Figure 2, which process tensors of a highly compressed spatial dimensions $H/8$ and $H/16$, respectively. Hence, for small input images, such layers cannot exploit the spatial capability of the convolutions.
%Hence, for 256x256 images, the hidden tensors addressed by those layers have a spatial size of $64\times 64$ and $32x32$ respectively, that are much larger than the corresponding tensor for 32x32 input images, these layers perform convolutional operations from a tensor having 4x4 spatial sizes, in which image's features are no longer well expressed.
%We believe this is duo to the the last convolutional layers, which, when the input dimension is $32\times 32$, they perform convolutional operations on a $764\times 4\times 4$ hidden-state,  which the behaviour is comparable to a fully connected layer.

\subsection{MAP Estimation}
This section evaluates
%an evaluation of the accuracy of the certifiable predictions for different values of $\varepsilon$, as well as 
the MAP estimation through the lower bound (LB) given by the UGNN discussed in Observation~\ref{oss:multi-motivation} and the other $1$-Lipschitz models.
\begin{figure}[h]
    \centering
    \includegraphics[clip, width=\columnwidth, trim=0 0.23cm 0 0.25cm]{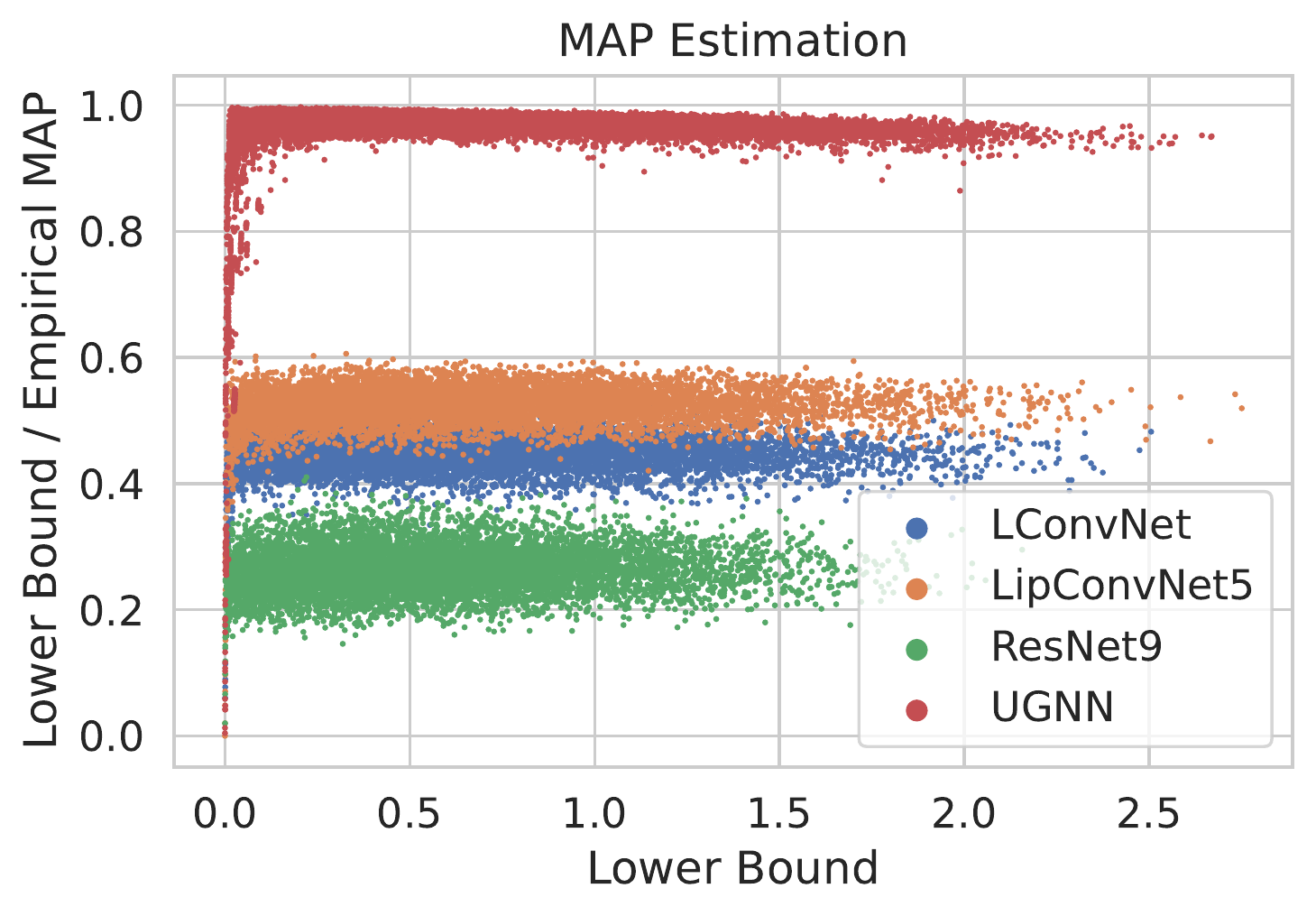}
    \caption{Evaluation of the lower-bound estimation of the MAP provided by the $1$-Lipschitz DNNs and the UGNN. The y-axis reports the ratio of the given lower bound and the MAP computed through an iterative-penalty algorithm.}
    \label{fig:scatter-IP-norm}
\end{figure}
%In the previous section we shown that if $f$ is an UGNN, then $\MM_f(x)$ is a lower bound of the MAP in $x$. Furthermore, such a $f$ satisfies (almost everywhere) the properties of a SDC, for which $\MM_f(x)$ coincides with the MAP in $x$ as discussed in Observation~\ref{oss:multi-motivation}.
%
Figure~\ref{fig:scatter-IP-norm} compares the ratio of the LB and the MAP between the $1$-Lipschitz DDNs and a UGNN with MaxMin and bounded upd, for the normalized inputs. The MAP is computed with the expensive Iterative Penalty procedure, as done in \cite{brau}. Note that our analysis considers the worst-case MAP, i.e., without \textit{box-constraints}, as also done by the compared $1$-Lipschitz models. Indeed, since image pixels are bounded in $[0,1]$, the MAP is itself a lower bound of the distance from the closest adversarial image. Table~\ref{tab:dist-table} reports statistics related to the LB/MAP ratio for different UGNNs, where the box-constrained (B.C.) MAPs were computed using the \add{Decoupling Direction Norm} strategy \cite{rony_2019}. The column \texttt{\#N} contains the number of samples correctly classified by the model and for which the MAP algorithm reached convergence. Note that, in all the tested cases, the LB provided by the UGNN resulted to be tighter than the other $1$-Lipschitz DNNs. Similar considerations hold for other model configurations (see Appendix).
\begin{table}[h]
    \centering
    \begin{tabular}{lccl}
\toprule
                  \textbf{Model} &  \textbf{LB/MAP} &  \textbf{\#N} & \textbf{B.C} \\
\midrule
                  %ResNet9 (norm) & $.21\!\pm\!.042$ &          7900 &       \cmark \\
                   ResNet9 (raw) & $.34\!\pm\!.063$ &          6669 &       \cmark \\
             %LargeConvNet (norm) & $.36\!\pm\!.036$ &          2148 &       \cmark \\
              %LipConvNet5 (norm) & $.41\!\pm\!.060$ &          7838 &       \cmark \\
              LargeConvNet (raw) & $.46\!\pm\!.057$ &          7219 &       \cmark \\
               LipConvNet5 (raw) & $.58\!\pm\!.069$ &          6911 &       \cmark \\
  %\textbf{UGNN+OPLU+orth (norm)} & $.67\!\pm\!.129$ &          7101 &       \cmark \\
   %\textbf{UGNN+OPLU+upd (norm)} & $.67\!\pm\!.131$ &          7281 &       \cmark \\
 %\textbf{UGNN+MaxMin+upd (norm)} & $.67\!\pm\!.130$ &          7311 &       \cmark \\
%\textbf{UGNN+MaxMin+orth (norm)} & $.69\!\pm\!.109$ &          4386 &       \cmark \\
    \textbf{UGNN+OPLU+updU (raw)} & ${\bf.70\!\pm\!.090}$ &      7125 &       \cmark \\
   \textbf{UGNN+OPLU+updB (raw)} & ${\bf.70\!\pm\!.093}$ &      7098 &       \cmark \\
 \textbf{UGNN+MaxMin+updB (raw)} & ${\bf.71\!\pm\!.087}$ &      7114 &       \cmark \\
  \textbf{UGNN+MaxMin+updU (raw)} & ${\bf.71\!\pm\!.088}$ &      7118 &       \cmark \\
  \midrule
                  ResNet9 (norm) & $.26\!\pm\!.036$ &          7904 &       \xmark \\
%                   ResNet9 (raw) & $.44\!\pm\!.056$ &          6663 &       \xmark \\
             LargeConvNet (norm) & $.44\!\pm\!.027$ &          7933 &       \xmark \\
              LipConvNet5 (norm) & $.52\!\pm\!.031$ &          7840 &       \xmark \\
%              LargeConvNet (raw) & $.58\!\pm\!.046$ &          2429 &       \xmark \\
%               LipConvNet5 (raw) & $.74\!\pm\!.049$ &          6912 &       \xmark \\
%    \textbf{UGNN+OPLU+upd (raw)} & ${\bf.93\!\pm\!.101}$ &          6755 &       \xmark \\
  \textbf{UGNN+OPLU+updB (norm)} & ${\bf.96\!\pm\!.046}$ &          7215 &       \xmark \\
   \textbf{UGNN+OPLU+updU (norm)} & ${\bf.96\!\pm\!.047}$ &          7282 &       \xmark \\
 \textbf{UGNN+MaxMin+updU (norm)} & ${\bf.96\!\pm\!.051}$ &          7316 &       \xmark \\
\textbf{UGNN+MaxMin+updB (norm)} & ${\bf.96\!\pm\!.044}$ &          7327 &       \xmark \\
\bottomrule
\end{tabular}

    \caption{Evaluation of the LB/MAP ratio deduced by the output of the models with/without Box Constraint.}
    \label{tab:dist-table}
\end{table}
\iffalse%
\begin{figure}[h]
  \begin{subfigure}{0.49\columnwidth}
  \includegraphics[width=\textwidth, keepaspectratio]{img/robust_lineplot-lipschitz.pdf}
  \caption{}
  %\caption{Goodness of the MAP estimation among the models computed with the IP strategy. Amount of samples are summarized in Table REF}
  \label{fig:robust-lipschitz}
  \end{subfigure}
  %
  \begin{subfigure}{0.49\columnwidth}
  \includegraphics[width=\textwidth, keepaspectratio]{img/robust_lineplot-ugnn_upd.pdf}
  \caption{}
  %\caption{MAP problem augmented with the box constraint computed with DDN strategy. Amount of samples are summarized in Table REF}
  \label{fig:robust-UGNN}
  \end{subfigure}
  \caption{Accuracy of the certifiable classifications w.r.t different values of $\varepsilon$.}
  \label{fig:rob-accuracy}
\end{figure}
\fi%

\subsection{Certifiable Robust Classification}
Figure~\ref{fig:rob-accuracy} shows a close comparison of \add{the accuracy of the (certifiable) $\varepsilon$-robust classifications} for different values of $\varepsilon$, i.e., the percentage of correctly classified samples with a LB lower than $\varepsilon$.
\begin{figure}[h]
    \centering
    \includegraphics[clip, width=\columnwidth, trim=0 0.23cm 0 0.25cm]{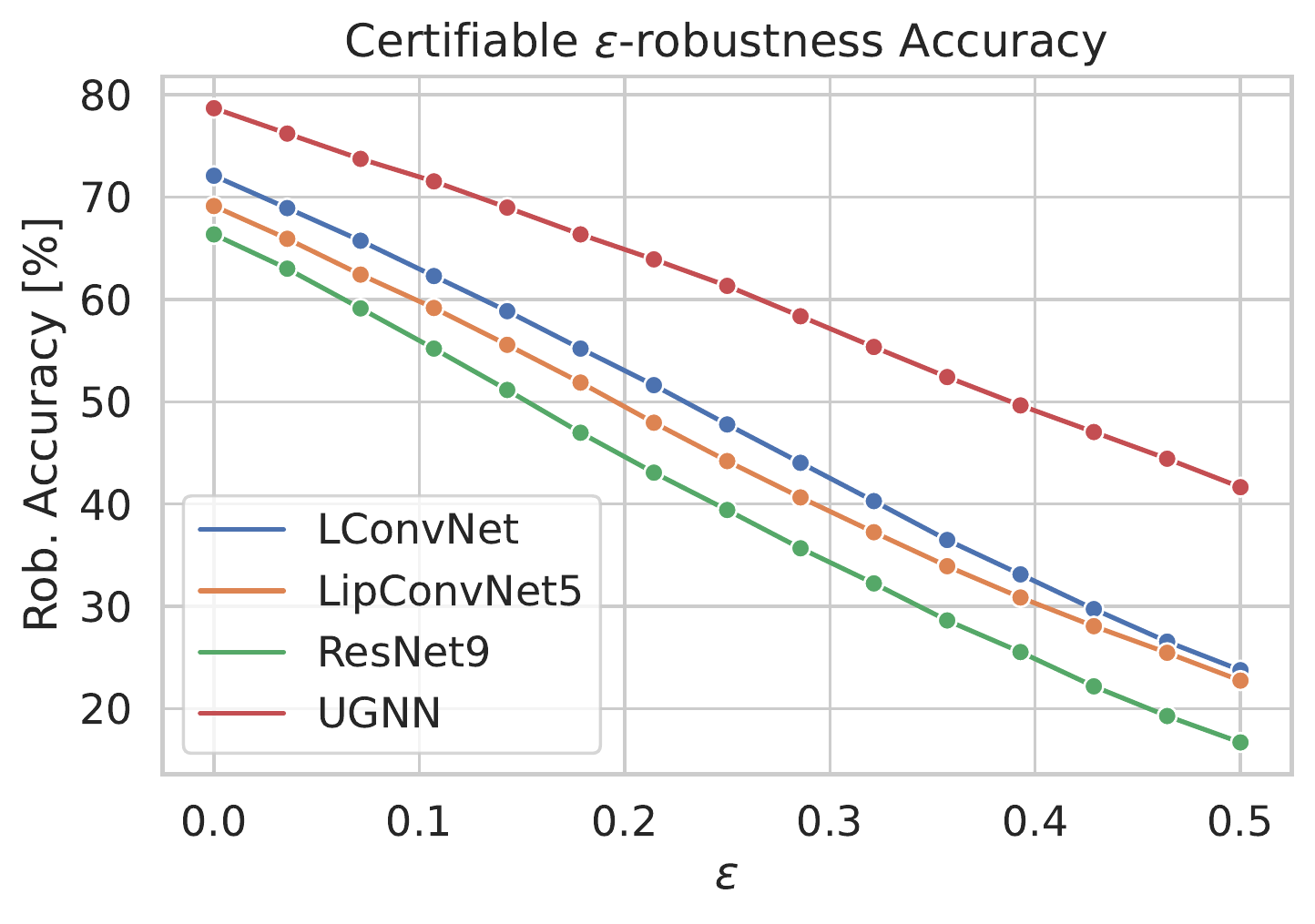}
    \caption{Accuracy of the certifiable $\epsilon$-robust classifications.}    
    \label{fig:rob-accuracy}
\end{figure}
We selected the UGNN with the highest accuracy (MaxMin-updB-256-raw). The tests for the 32x32 input size are provided in Appendix. The $1$-Lipschitz models were trained on raw inputs, where the best run has been selected. For these models, to handle 256x256 images, an initial nearest interpolation from $256$ to $32$ is applied. This transformation is necessary since, differently from the UGNN, they are not adaptive to different input sizes. Note that the interpolation preserves both the accuracy and the $1$-Lipschitz property.
As shown in Fig.~\ref{fig:rob-accuracy}, the UGNN outperforms the other models for all the tested $\varepsilon$ values.
%The certifiable accuracy of the $1$-Lipschitz models resulted to be comparable with all the tested UGNNs.
\section{Conclusion}
This paper presented a novel family of classifiers, named \textit{Signed-Distance Classifiers} (SDCs), which
%, in addition to infer the class of a given input, 
provides the \textit{minimal adversarial perturbation} (MAP) by just computing the difference between the two highest output components, thus offering an online-certifiable prediction.
%Hence, such classifiers can provide certifiable predictions at run time.
%Furthermore, we deduced a condition that characterized an SDC by extending a known result on the Signed Distance Functions.

To practically implement an SDC, we developed a novel architecture, named \textit{Unitary-Gradient Neural Network} (UGNN), which satisfies (almost-everywhere) the characterization property of an SDC.
To design this model, we proposed a new fully-connected layer, named Unitary Pair Difference (UPD), which features unbounded weight matrix while preserving the unitary-gradient property.

%the aforementioned characterization property in order to behave like an SDC as close as possible. To craft such a model, we proposed a new fully-connected layer, named UPD, which features unbounded weight matrix.
%Several experiments were conducted to show the advantages of the proposed model and compare it with the most related certifiable $1$-Lipschitz models. The experiments highlighted the limits of a UGNN in terms of accuracy and showed a comparable robustness with respect to the other DNNs.
Several experiments were conducted to compare the proposed architecture with the most related certifiable $1$-Lipschitz models from previous work.
The experiments highlighted the performance of the UGNN in terms of accuracy, certifiable robustness, and estimation of the MAP, showing promising results.

Future work will focus on improving the UGNN.
Furthermore, as pointed out by other authors, additional investigations are needed to tackle practical open problems in this field, such as addressing dataset with many classes and improving learning strategies. 
%Finally, since the experiments showed a high sensitivity of the MAP estimation from the activation function, future works will be focused on deducing empirical and theoretical investigations. Furthermore, as experienced by other authors, we got difficulties in training the models on data-sets with many classes, hence, future works will investigate this aspect.

\bibliography{biblio}
\clearpage
\appendix
\twocolumn[
  \begin{@twocolumnfalse}
\begin{center}
  \Large
  \textbf{Technical Appendix of ``Robust-by-Design Classification via Unitary-Gradient Neural Networks''}
\end{center}
\vspace{1pt}
%\begin{center}
%  \Large
%  \textbf{Anonymous submission}
%\end{center}
\vspace{10pt}
  \end{@twocolumnfalse}
]
%\printglossary[type=\acronymtype]
%\printglossary
\section{Gradient Norm Preserving\\ Activation Functions}
%\subsection{Characterization of the Component-Wise GNP Activation Functions}
%\label{app:activation-function}
Component-wise activation functions that satisfy
Property~\ref{eq:orth-constr} can be completely characterized; this is the
aim of the following lemma.
\begin{lem}[GNP Component-wise Activation Functions]
  The only component-wise activation functions that guarantee the orthogonal
  property~\ref{eq:orth-constr} are the \textit{piecewise-linear} functions with slope
  $1$ or $-1$.
  \begin{proof}
    Let $h:\R\to\R$ a scalar function, and let $\textbf{h}:\R^n\to\R^n$ the
    tensor-wise version of $h$ defined as $\textbf{h}(x) = (h(x_i))_i$.
    Observe the corresponding Jacobian matrix is always represented by a diagonal matrix 
    $\jac(\textbf{h})(y)=\diag(h'(y_1),\dots,h'(y_n))$. 
    The orthogonal condition on the Jacobian rows is only guaranteed if $h(y)$
    solves the differential equation 
    \begin{equation}
      \label{eq:hdiff}
      (h'(y))^2=1
    \end{equation}
    Observe that all the solutions of Equation~\ref{eq:hdiff}
    are of the form $h(x) = \sum_{i=1}^P (\alpha_i x +\beta_i)\1_{U_i}(x)$ 
    where $\alpha_i\in\{1,-1\}$, $\beta_i\in\R$, 
    and $\{U_i\}_i$ is a discrete partition of $\R$.
    Observe in conclusion that $h(x)=|x|$ solves Equation~\ref{eq:hdiff}.
  \end{proof}
\end{lem}
\subsection{Tensor-wise GNP activation functions}
The OPLU activation function was introduced in \cite{chernodub} and
recently generalized from \cite{anil_fc}. Accordingly with the original paper, we assume the following definition.
\begin{defn}[OPLU]
  The 2-dimensional version is defined as follows
  \begin{equation}
    \begin{aligned}
      \oplu_1&:\R^2\to\R^2\\
      &
      \begin{bmatrix}
        x\\
        y
      \end{bmatrix}\mapsto
      \begin{bmatrix}
        \max (x,y)\\
        \min (x,y)
      \end{bmatrix}.
    \end{aligned}
  \end{equation}
  The generalization to higher dimensional spaces is the following
  \begin{equation}
    \begin{aligned}
      \oplu_n&:\R^{2n}\to\R^{2n}\\
      &
      \begin{bmatrix}
        x_1\\
        \vdots\\
        x_n
      \end{bmatrix}\mapsto
      \begin{bmatrix}
        \oplu_1(x_1)\\
        \vdots\\
        \oplu_1(x_n)\\
      \end{bmatrix}.
    \end{aligned}
  \end{equation}
\end{defn}
\section{Characterization of the Signed Distance Functions}
This section contains a proof of Theorem~\ref{thm:sdf-characterization}.
For the sake of a clear comprehension, before providing the proof, let us remind some classical results. The following theorems are known as Existence and Uniqueness of Solutions of Ordinary Differential Equations (ODE) and Implicit Function Theorem.
\begin{thm}[Existence and Uniqueness of ODE solutions]
\label{thm:ode}
Let $\UU\subseteq\R^n$ be an open subset of $\R^n$, and let $F:\UU\to\R^n$ a smooth function, i.e., $F\in C^\infty(\UU)$, then the following statements hold.
\begin{enumerate}[label=\roman*)]
    \item For each $t_0\in\R$ and $x_0\in\UU$, there exists $I_0\subseteq\R$ and $\UU_0\subseteq\UU$ open sets, with $(t_0,x_0)\in I_0\times \UU_0$, such that for each $x\in\UU_0$ there exists a solution $u_x:I_0\to\UU$ of the following Cauchy-problem
    \begin{equation}
        \begin{cases}
            \dot u (t)= F(u(t))\\
            u(0) = x;
        \end{cases}
        \label{eq:cauchy}
    \end{equation}
    where we keep the notation $u_x$ to highlight that $x$ is the starting point of the solution of Problem~\ref{eq:cauchy}.
    \item The map $\Theta: I_0 \times \UU_0\subseteq\UU$, namely \emph{flux}, defined by $\Theta(t,x):=u_x(t)$, is in $C^\infty$;
    \item If $u_x, v_x$ are two solutions of Equation~\eqref{eq:cauchy}, then $u_x\equiv v_x$;
\end{enumerate}
\begin{proof}
    Refer to \cite[pp.66-88]{lang}.
\end{proof}
\end{thm}

The implicit function theorem can instead be stated as follows.
\begin{thm}[Implicit Function Theorem (Dini)]
    \label{thm:implicit}
    Let $G:\R\times \UU\to\R$ be a smooth function defined on an open set $\R\times\UU$. If $p\in\UU$ is such that 
    \[
        G(0,p)=0\quad\mbox{and}\quad \frac{d G}{d t}(0,p)\ne 0,
    \]
    then there exists an open set $\Omega\subseteq\UU$ and a smooth function $\varphi:\Omega\to\R$ such that
    \begin{equation}
        \forall x\in\Omega,\quad G(\varphi(x),x)=0.
    \end{equation}
\begin{proof}
    The proof can be deduced by \cite[Thm. 5.9]{lang}, where $V:=\R$,  $U:=\UU$, $U_0:=\Omega$ and $(a,b)=(p,0)$. 
\end{proof}
\end{thm}
Finally we can leverage these results to prove the main theorem of the paper. 

\smallskip
\noindent
\textbf{Theorem~\ref{thm:sdf-characterization}}
%\begin{thm}
\emph{
  Let $\UU\subseteq\R^n$ be an open set, 
  and let $f:\R^n\to\R$ be a function, smooth in $\UU$, such that $\BB_f\subseteq\UU$. If $f$ has a unitary gradient in $\UU$, then 
  there exists an open set $\Omega_f\subseteq\UU$ such that $f$ coincides in $\Omega_f$ with the signed distance function from $\BB_f$. 
  Formally,}
  \begin{equation}
      \|\nabla f_{\restriction\UU} \| \equiv 1 \quad \Rightarrow \quad\exists\Omega_f\subseteq\UU,\quad  f_{\restriction \Omega_f} \equiv d^*_{f\restriction \Omega_f}.
      \label{eq:unitary-constraint}
  \end{equation}
%  \end{thm}
\begin{proof}
We have to prove that there exists an open set $\Omega_f\subseteq\UU$ such that the unitary gradient property in $\UU$ implies that $f(x)= d_f^*(x)$ for all $x\in\Omega_f$. The proof is divided in two main parts:
\begin{enumerate}[label=(\roman*)]
    \item Let us consider the following ordinary differential equation with initial condition (a.k.a. Cauchy problem)
\begin{equation}
  \begin{cases}
    \dot u(t) = \nabla f(u(t))\\
    u(0) = x
  \end{cases}
  \label{eq:gradient-flux}
\end{equation} where $x\in\UU$. We show that there exists an open set
        $\Omega_f\supseteq\BB_f$ such that each $x \in \Omega_f$ can be reached by a solution $u_p$ of the Cauchy-problem~\eqref{eq:gradient-flux}, i.e., $\exists s\in\R$ such that $x=\Theta(s,p) := u_p(s)$ for some $p\in\BB_f$;
    \item We show that any trajectory of the
        flux corresponds the minimal geodetic (i.e., the shortest path) between the hyper-surfaces of the form
        $f^{-1}(t)$ and $\BB_f$. This can be obtained by explicitly deducing a close form of $f$ on $\Omega_f$.
\end{enumerate}

Let us start with the existence of such a $\Omega_f$. Since $f$ is smooth, then $F:=\nabla f$ satisfies the hypothesis of Theorem~\ref{thm:ode}, by which we can deduce that for each $p\in\BB_f$ there exists an open set $I_p\times U_p\subseteq\R\times\UU$ such that the flux
\begin{equation}
    \begin{aligned}
        \Theta&:I_p\times\UU_p\to\UU\\
        &\quad(t,x)\mapsto u_x(t)
    \end{aligned}
\end{equation}
is of class $C^\infty$ (where remember that $u_x$ is the solution of the ODE~\eqref{eq:cauchy} with starting point in $x$). Let $G:I_p\times \UU_p\to\R$ be the smooth function defined by $G(t,x):=f(\Theta(t,x))$. 
By~\eqref{eq:gradient-flux},  $\frac{d \Theta}{d t} (0,p) = \dot u_p(0) = \nabla f(u_p(0))$ and $\Theta(0, p) = u_p(0)=p$, hence it is possible to observe that
\begin{equation}
        G(0,p)=f(\Theta(0,p))=f(p)=0
\end{equation}
and that
\begin{equation}
        \frac{d G}{d t} (0,p) =\nabla f(p)^T\frac{d \Theta}{d t} (0,p) =
        \nabla f(p)^T\nabla f(p).
\end{equation}
We then deduce by the Implicit Function Theorem~\ref{thm:implicit} that there exists an open set $\Omega_p\subseteq \UU_p$ such that
\begin{equation}
    \forall x\in\Omega_p,\quad \exists t\in I_p\quad:\quad G(t,x)=0,
\end{equation}
from which
\begin{equation}
    \forall x\in\Omega_p,\quad \exists t\in I_p\quad:\quad u_x(t)\in\BB_f.
\end{equation}
From the uniqueness of the solution stated in Theorem~\ref{thm:ode}, this implies that, for each $x\in\Omega_p$, there exists $q\in\Omega_p\cap\BB_f$ and an instant $t\in I_p$ such that $u_q(t)=x$. Finally, by considering $\Omega_f:=\cup_{p\in\BB_f} \Omega_p$, the first step of the proof is concluded.

Now, we want to prove that the trajectory of the dynamic system coincides with the geodetic (the curve of minimal length) from any $x\in\Omega_f$ and for any $B_p$.
Let $u_p:I_p\to\Omega_f$ be the solution of \eqref{eq:cauchy} with starting point in $p\in\BB_f$, and let $x=u_p(s)$ be the point of the trajectory for $s\in I_p$. Let us consider a function $\gamma(t):=u_p(ts)$ of the form $[0,1]\to\Omega_f$ to denote the curve that connects $p$ and $x$. Observe that the length of $\gamma$ can be found by considering the following formula
\begin{equation}
    L(\gamma) := \int_{0}^1 \|\dot\gamma(t)\|\,dt = \int_0^1 |s|\|\dot u_p(t)\|\, dt.
\end{equation}
Since $\|\dot u_p\| = \|\nabla f(u(t))\| = 1$ we can deduce that the length of $\gamma$ is $L(\gamma)=|s|$.

Let $\zeta:[0,1]\to\UU$ be any other curve that connects $p$ and $x$. Observe that the following chain of inequalities holds

\begin{equation}
    \begin{aligned}
        L(\zeta)&=\int_0^1 \|\dot\zeta\|\,dt \ge\\
        &\ge \int_0^1 \left\vert\left\langle\dot \zeta(t),\nabla f(\zeta(t))\right\rangle\right\vert\,dt \ge \\
    &\ge \left\vert \int_0^1 \frac{d}{dt} f(\zeta(t))\,dt\right\vert = |f(p) - f(x)| = |f(x)|,
    \end{aligned}
\end{equation}
where the first inequality is a direct consequence of the Cauchy-Schwarzt inequality ($\forall u,v\in\R^n,\,|\langle u,v\rangle| \le \|u\|\|v\|$).

It remains to prove that $L(\zeta)\ge L(\gamma)$. To do so, let us consider the following observation.

\begin{oss}
  If $p\in\BB_f$ and $s\in I_p$, then $f(u_p(s)) = s$.
  \begin{proof}
    Let $\varphi(s) = f(u_p(s))$ be the value of $f$ on the trajectory of the flux. Since $\dot\varphi = \langle\dot u_p(s), \nabla f(u_p(s))\rangle = 1$ we deduce $\varphi(s) = s + \varphi(0) = s$.
  \end{proof}
\end{oss}

This concludes the second step of the proof, since for each curve $\zeta$ that connects $p$ and $x$ we have that 
\[
L(\zeta) \ge |f(x)| = |s| = L(\gamma),
\]
hence $\gamma$ is the shortest path between $p$ and $x$, from which $|f(x)|= d_f(x)$.

In conclusion, the theorem is proved by observing that, for each $x\in\Omega_f$, there exists $p \in \BB_f$ such that $x = u_p(s)$ for some $s$. Indeed, by the definition of $\Omega_f$, let $q$ such that $x\in\Omega_q$, then there exists a $p\in\Omega_q\cap\BB_f$ such that $x =u_p(s)$ for some $s$.
\end{proof}

\subsection{An example of non-affine Signed Distance Function}
In the main paper, we observed that $f(x):=\|x\|-1$ is an instance of a non-affine signed distance function. Indeed, observe that, for each $x\in\R^n\setminus \{0\}$, the gradient of $f$ has unitary euclidean norm and it has the following explicit formulation $\nabla f(x) = \frac{x}{\|x\|}$. Furthermore, the minimal adversarial perturbation problem relative to $f$ can be written as follows
\begin{equation}
\begin{aligned}
  \min_{p\in\R^n} &\quad \| x- p\|\\
  \mbox{s.t.} &\quad p_1^2+\cdots+p_n^2 = 1
\end{aligned},
\end{equation}
and has a minimal solution of the form $x^*=\frac{x}{\|x\|}$. This fact can be proved by considering the associated Lagrangian function $\LL(p,\lambda):=\|x-p\|-\lambda (\|p\|^2-1)$, from which we can deduce that $p^*:=\frac{x}{\|x\|}$, is a stationary point of $\LL$, i.e., there exists a Lagrangian multiplier $\lambda^*$ such that $\nabla \LL(x^*,\lambda^*)=0$, realized for $\lambda^*=\pm \frac{1}{2}$.
\section{Extension to Multi Class Signed Classifiers}
This section contains the technical details for the proof of Observation~\ref{oss:multi-motivation} related to the definition of the signed distance classifier for multi-class classification. Let us first consider the following lemma that shows that Problem~\ref{eq:map} can be solved by
considering the smallest solution of a sequence of a minimum problems
\begin{lem}
  \label{lem:min-seq}
  Let $x\in\R^n$ classified from $f$ with the class $l$, $\hat k(x) = l$. Let, for each $j\ne l$, $g_j = (f_l-f_j)$, then 
  \[
    d_f(x) = \min_{j\ne l} d_{g_j}
  \] where $d_{g_j}$ is the solution of the Problem~\ref{eq:minimal-adversarial-problem} relative to the binary classifier $g_j$. In formulas,
 \begin{equation}
    \begin{aligned}
    d_{g_j}(x):= &\inf_{p\in\R^n} & \|p-x\|\\
           & \mbox{s.t.}    & f_l(p) - f_j(p)=0   
  \end{aligned}     
 \end{equation}
  \begin{proof}
    The main idea is to separately prove
    the two inequalities
    \begin{equation}
      \min_{j\ne l} d_{g_j}(x) \le d_f(x) \le \min_{j\ne l} d_{g_j}(x).
    \end{equation}
    The inequality on the right can be deduced by observing that, for each $j$, the solution $x^*_j$ of the Problem~\ref{eq:minimal-adversarial-problem}, relative to the function $g_{j}$, satisfies the constraints of the minimum problem~\ref{eq:map} relative to the function $f$. Hence, by the definition of minimum $d_f(x)=\|x-x^*\|\le \|x-x_j^*\|$. 
    
    The inequality on the left is deduced by observing that if $x^*$ is the solution of $d_f$ and if $j^*$ is such that $f_{j^*}=\max_{j\ne l} f_j(x^*)$, then $x^*$ satisfies the constraints of the Problem~\ref{eq:minimal-adversarial-problem} for the function $d_{g_{j^*}}$. Hence, 
    \[
      \min_{j\ne l} d_{g_j}\le d_{g_{j^*}} \le \| x - x^*\|,
    \]
    which concludes the proof.
  \end{proof}
\end{lem}

\noindent
\textbf{Observation 2.}  Let $f:\R^n\to\R^C$ a signed distance classifier and let $x\in\R^n$ a sample classified as $l = \hat k(x)$, and let $s:=\argmax_{j\ne l} f_j(x)$ the second highest component. Then, the classifier $f$
  \begin{enumerate}[label=(\roman*)]
  \item provides a fast way to certificate the
    robustness of $x$. In fact, $f_l(x) - f_s(x) = d_f(x)$, where $d_f(x)$ is the minimal adversarial perturbation defined by the Problem~\ref{eq:map}.
  \item provides the closest adversarial example to $x$, being 
    \[
      x^* = x - (f_l(x)-f_s(x)) \nabla (f_l-f_s)(x)
    \]
    where $x^*$ is the unique solution of Problem~\ref{eq:map} in $x$.
 \end{enumerate}
 \begin{proof}
    The first statement is a direct consequence of Lemma~\ref{lem:min-seq}. Consider the following chain of equalities 
   \begin{equation}
     \begin{aligned}
       d_f(x) &= \min_{j\ne l} d_{g_j}(x) = \min_{j\ne l} (f_l -f_j)(x)\\
       &= f_l - \max_{j\ne l} f_j(x) = (f_l-f_s)(x).\\
     \end{aligned}
   \end{equation}
   where the second equivalence is given by the definition of a signed distance classifier. The second statement is a consequence of Observation~\ref{oss:sign-dist-grad}, indeed, $\nabla (f_l-f_s)(x)$ provides the direction of the shortest path to reach $B_{ls}$.
 \end{proof}

\iffalse
\begin{thm}
  \label{thm:sufficient}
  Let $\UU\subseteq\R^n$ a connected open set. If $f:\UU\to\R^C$ is such that 
  \begin{equation}
    \jac f(x)\jac f(x)^T=\frac{a^2}{2}Id,
    \label{eq:jacobian-condition}
    \tag{CGN}
  \end{equation}
  then $\frac{1}{a}f$ is multi signed distance neural network.
  \begin{proof}
    Observe that the condition in \cref{eq:jacobian-condition} implies that
    $\|\nabla f_i\|^2\equiv\frac{a^2}{2}$ and that $\nabla f_i^T\nabla
    f_j\equiv0$ for each couple $i,j$. We need to prove that
    $\frac{1}{a}(f_i-f_j)$ is a
    signed distance function for each couple $i,j$. Observing that
    \begin{equation}
      \|\nabla f_i-\nabla f_j\|^2\equiv\underbrace{\|\nabla
      f_i\|^2}_{a^2/2}+\underbrace{\|\nabla f_j\|^2}_{a^2/2}-2 \underbrace{\nabla
      f_i^T\nabla f_j}_{0} \equiv a^2,
    \end{equation}
    and by applying the characterization in \Cref{thm:sdf-characterization}, we deduce
    the thesis.
  \end{proof}
\end{thm}
\fi
\section{The \texttt{PixelUnshuffle} is a gradient norm preserving layer}
Pixel-Unshuffle layer has a fundamental role in crafting a unitary gradient neural network, since it allows increasing the number of channels through the internal activations and, simultaneously, keeping the GNP property of the convolutions. A Pixel-Unshuffle layer, with scaling-size of $r$, transforms an input $x\in\R^{C\times rH\times rW}$ only by rearranging its entries to provide an output tensor of shape $r^2C\times H\times W$. Such a layer, can be discribed as the inverse of the \emph{pixel-shuffle} layer $S$ described as follows
\[
    (S x )\left[c,i,j\right] = x\left[rC\,x\%r+C\,y\%r+c,\left\lfloor\frac{i}{r} \right\rfloor,\left\lfloor\frac{i}{r} \right\rfloor\right],
\]
where $[c,i,j]$ means the entry $i,j$ of the $c$-th channel of the leftmost tensor. Hence, observe that the vectorized version of S can be described as a map $\hat S :\R^{m}\to\R^{m}$ such that $(\hat S (x))_i = x_{\sigma(i)}$, where $m = r^2CHW$ and $\sigma:\{1,\cdots,m\}\to\{1,\cdots,m\}$ is a one-to-one permutation map. Finally observe that, 
\begin{equation}
    \jac (\hat S)_{ij} =
    \begin{cases}
        1 &j=\sigma(i)\\
        0 &\mbox{otherwise,}
    \end{cases}
\end{equation}

from which we can deduce that each row of the Jacobian contains one and only one not-zero entry (that is $1$), and, that every two rows are different. This very last statement directly implies the orthogonality of $\jac(\hat S)$ in each $x$. In conclusion the GNP property of the pixel-unshuffle layer $S^{-1}$, can be deduced by observing that, if $\hat S^{-1}$ is the vectorized version of $S^{-1}$, then
\[
\begin{split}
\jac(\hat S^{-1}) \jac(\hat S^{-1})^T = (\jac (\hat S)^{-1})(\jac(\hat S)^T)^{-1}= \\
= (\jac(\hat S)^T\jac(\hat S))^{-1} = I^{-1} = I.
\end{split}
\]

\section{Parameterized Unitary Pair Difference Layer}
This section aims at describing how the objective function $\Psi$ can be efficiently computed by exploiting the parallelism. Let us consider the family of matrices $A^{(k)}$, within $\binom{k}{2}$ rows and $k$ columns, recursively defined as follows
\begin{equation}
    A^{(2)} = \left(\begin{matrix}
      1 & -1
    \end{matrix}\right),\quad
    A^{(k)}=
    \left(\begin{array}{@{}c|c@{}}
        \begin{matrix}
            1\\
            \vdots\\
            1\\
            \\
        \end{matrix}
        & -Id_{k-1} \\
        \hline
        \begin{matrix}
        \\
            0\\
            \vdots\\
            0
        \end{matrix}
        &
        \begin{matrix}
            A^{(k-1)}
        \end{matrix}
    \end{array}\right)
      \quad \forall k\ge 2.
\end{equation}

Hence, observe that if $U\in\R^{C\times m}$ is some matrix, then the resulting matrix product $A^{(C)} \, U$ corresponds to a matrix where each row is one of the possible difference between two rows of $U$. In formulas
\begin{equation}
    A^{(C)} U = A^{(C)} \left( 
    \begin{matrix}
    U_1^T\\
    \vdots\\
    U_C^T
    \end{matrix}
    \right) 
    =
    \left(
    \begin{matrix}
    (U_1 - U_2)^T\\
    \vdots\\
    (U_1-U_C)^T\\
    (U_2-U_3)^T\\
    \vdots
    \end{matrix}
    \right).
\end{equation}
This allows exploiting the parallelism of the GPUs in order to efficiently compute the objective function $\Psi$.
\begin{figure}[ht]
    \centering
    \includegraphics[width=\columnwidth]{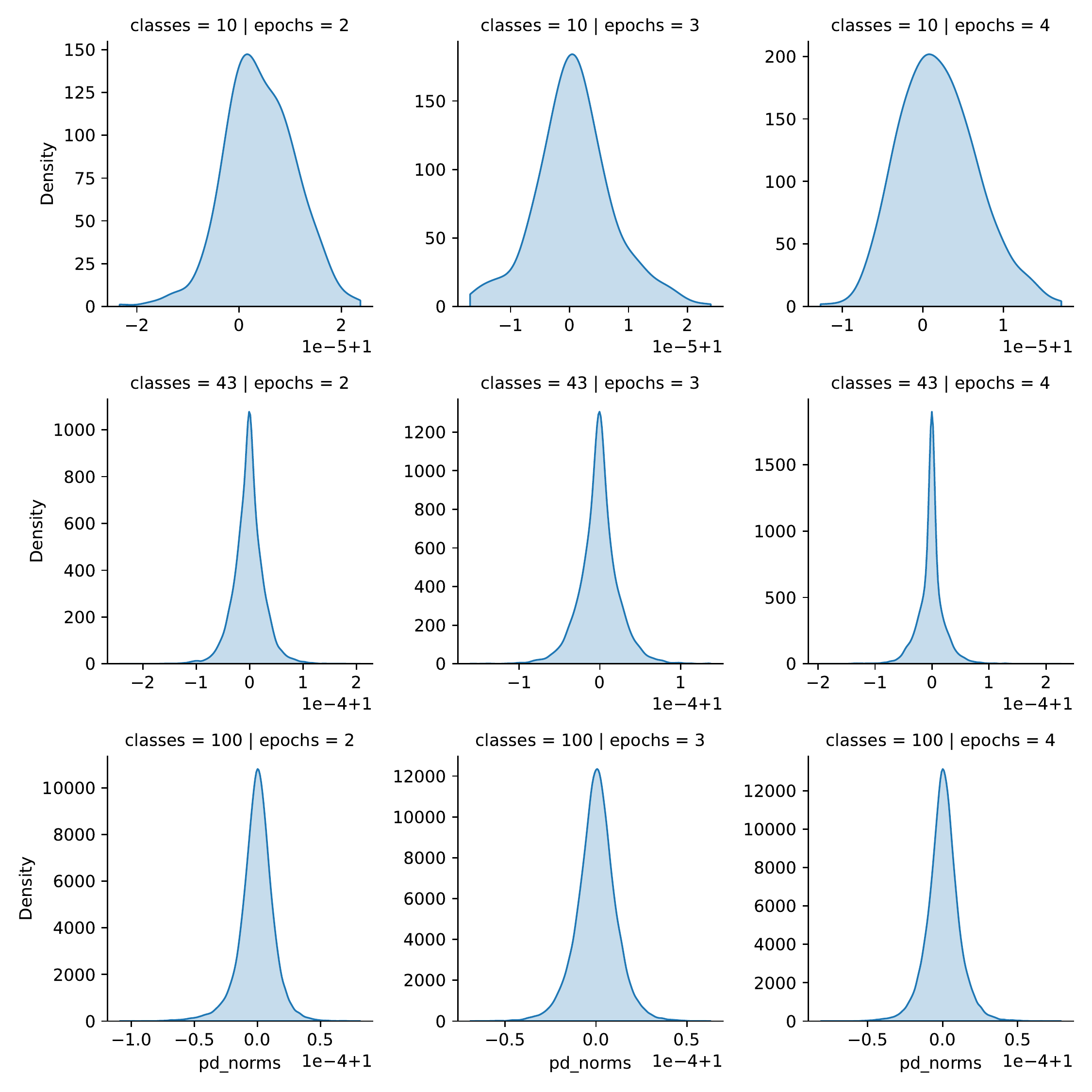}
    \caption{Distribution of the pair row differences of weight matrices $W=\texttt{UPD}(U)$ obtained by applying the L-BFGS for $2,3,4$ steps. The analysis involves matrices with $10,43,100$ rows and $512$ columns. Distributions are computed by evaluating the euclidean norm of all the pair-wise difference of the rows of the matrix $W$ for $10$ random generated parameters $U$.}
    \label{fig:distplot_upd}
\end{figure}
In conclusion, experimental tests reported in Figure~\ref{fig:distplot_upd} show that $3$ iterations of the L-BFGS algorithm are sufficient to obtain a UPD matrix $W$ whose differences between rows have an euclidean norm in the range $1 \pm 10^{-5}$ for the case of interest ($C=10$).
\section{Unitary Gradient Neural Network}
\subsection{The Unitary Gradient Property}
\begin{figure*}[ht]
    \centering
    \includegraphics[width=\textwidth]{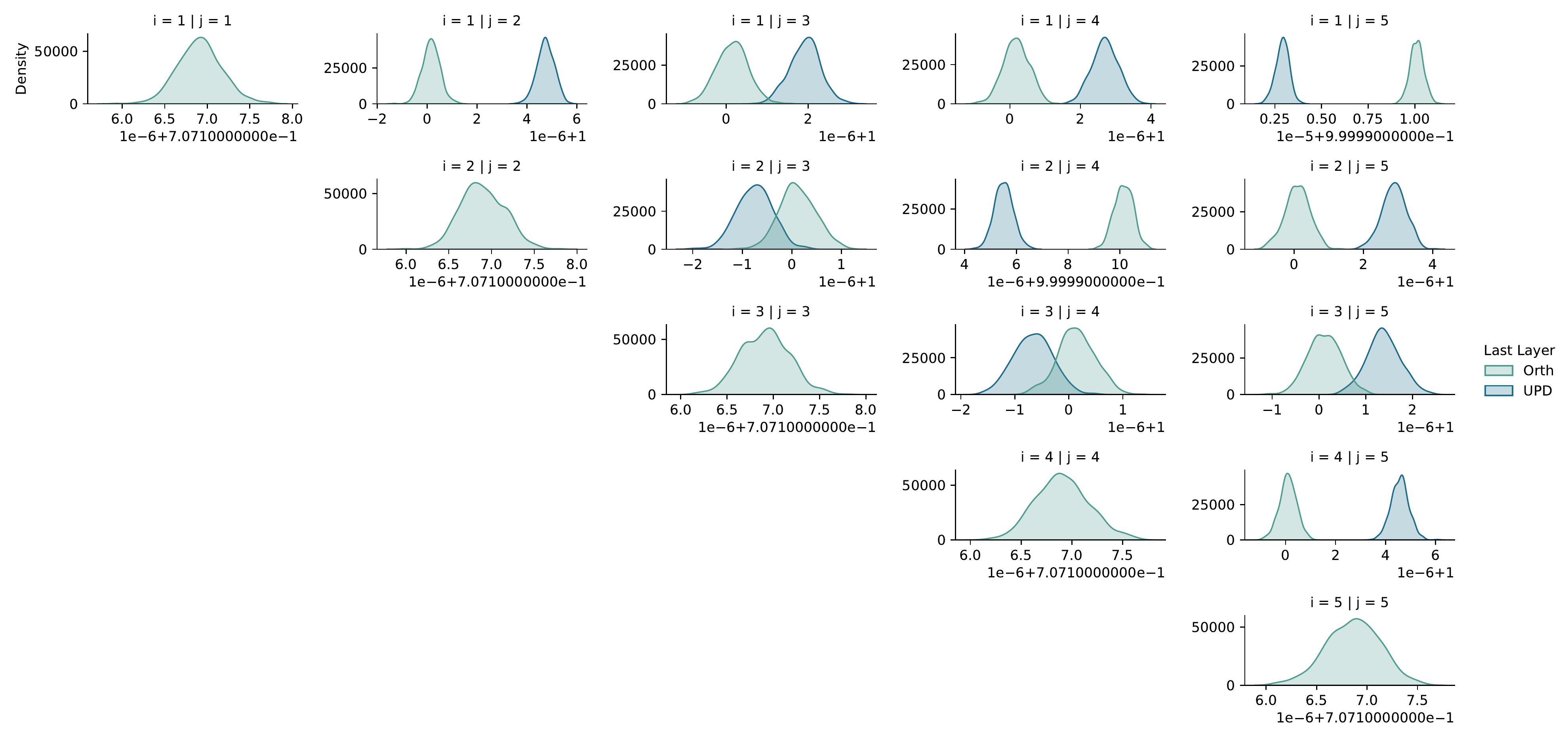}
    \caption{Empirical evaluation of the unitary gradient property for the UGNN model within $5$ output classes. The plot in the coordinate $i,\,j$ represents the distribution of the euclidean norm of $f_i - f_j$ for $512$ random samples with Gaussian distribution.}
    \label{fig:distplot_unitary_grad}
\end{figure*}

This section aims at empirically evaluating the euclidean norm of the pair difference $f_h -f_k$ to show that $\|\nabla(f_h-f_k)\|$ is numerically equal to $1$. Distribution plots in Figure~\ref{fig:distplot_unitary_grad} show the distribution of the norm of the difference $f_h-f_k$ for a classifier with $5$ output classes.
\subsection{Certifiable Robust Classification through UGNNs}
This section aims at providing further details related to the robustness statements in Observation~\ref{oss:robustness} reported below.
Before going deeper in the details, it is worth to remind a known results.
\begin{lem}\label{lemma:1lip-unitary}
    Let $f:\R^n\to\R$ a continuous function, differentiable almost-everywhere. If $\|\nabla f(x)\|=1$ in each differentiable point $x$, then $f$ is $1$-Lipschitz.
\begin{proof}
Let $x,y\in\R^n$ and $\phi(t):=x+(y-x)t$ be the straight line that connects the two points. Observe that, by the hypothesis, the function $(f\circ \phi)'$ is continuous almost everywhere, from which
\[
    f(y) - f(x) = \int_0^1 (f\circ \phi)'(t)\,\mbox{dt}=\int_0^1 \nabla f(\phi(t))^T(y-x)\,\mbox{dt}.
\]
By applying the absolute value and considering the Cauchy–Schwarz inequality ($\forall v,w,\,\left\langle v,w\right\rangle\le\|v\|\cdot\|w\|$), the following inequality holds
\[
\begin{split}
    \vert f(y) - f(x)\vert = \left\vert\int_0^1 \nabla f(\phi(t))^T(y-x)\,\mbox{dt} \right\vert\le\\ \le\int_0^1 \|\nabla f(\phi(t))\| \|y-x\|\,\mbox{dt} = \|x-y\|,
\end{split}
\]
from which the thesis follows.
\end{proof}
\end{lem}
Finally, the observation can be easily proved.

\noindent
\textbf{Observation 5} (Certifiable Robustness). 
If $f$ is a UGNN, then $\hat k$ is $\MM_f(x)-$robust in $x$. In other words, $\MM_f(x)$ is directly a lower bound of the MAP in $x$.
\begin{proof}
Let us assume $x\in\R^n$ such that $\nabla f(x)$ is defined and such that $\hat k(x)=l$. Let $\MM_f(x)$ defined as follows
\[
    \MM_f(x):=f_l(x) - \max_{j\ne l} f_j(x)= \min_{j\ne l} (f_l-f_j)(x)
\]
By the definition of UGNN, observe that $\|\nabla (f_l -f_j)\|\equiv 1$, for each $j\ne l$, hence, by Lemma~\ref{lemma:1lip-unitary}, $f_l-f_j$ is $1$-Lipschitz. By the definition of $1$-Lipschitz functions, we deduce that, for each $\delta$ such that $\|\delta\| < \MM_f(x)$,
\[
    \forall j\ne l,\quad \vert(f_l-f_j)(x+\delta) - (f_l-f_j)(x)\vert < \|\delta\| \le  \MM_f(x)
\]
where the first inequality is due to the lipschitz property and the second is due to the choose of $\delta$. By considering only the negative part of the absolute value, we then obtain that
\[
    \forall j\ne l, \quad (f_l-f_j)(x+\delta) > (f_l-f_j)(x) - \MM_f(x) \ge 0.
\]
This implies that $f_l(x+\delta) > f_j(x+\delta)$ for all $j\ne l$ from which we can deduce
that $\hat k(x+\delta) = l$.
In conclusion, let $\|\delta^*\|$ the minimal adversarial perturbation in $x$, then since 
\[
   \forall \delta,\qquad \hat k(x) \ne \hat k(x+\delta) \quad\Rightarrow\quad \|\delta\|\ge \MM_f(x),
\]
then by considering the inferior on $\|\delta\|$, we obtain that
\[
    \|\delta^*\| = \inf\{\|\delta\|\,:\, \hat k(x) \ne \hat k(x+\delta)\} \quad \Rightarrow \quad \|\delta^*\|\ge \MM_f(x),
\]
which concludes the proof.
\end{proof}
\section{Supplementary Experimental Material}
\subsection{Further MAP Estimations Analysis}
\begin{figure*}[ht]
  \begin{subfigure}{\columnwidth}
    \includegraphics[width=\textwidth, keepaspectratio]{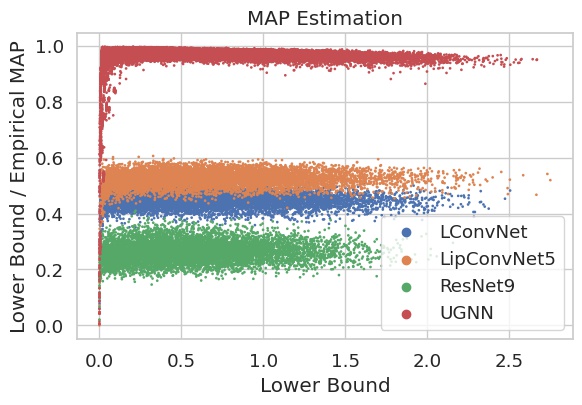}
    \subcaption{Normalized Inputs. UGNN with OPLU and upd.}
  \end{subfigure}%\hfill
  \begin{subfigure}{\columnwidth}
    \includegraphics[width=\textwidth, keepaspectratio]{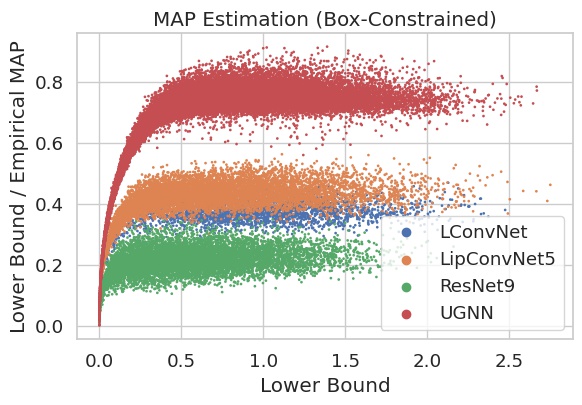}
    \subcaption{Normalized Inputs.  UGNN with OPLU and upd.}
  \end{subfigure}
  \begin{subfigure}{\columnwidth}
    \includegraphics[width=\textwidth, keepaspectratio]{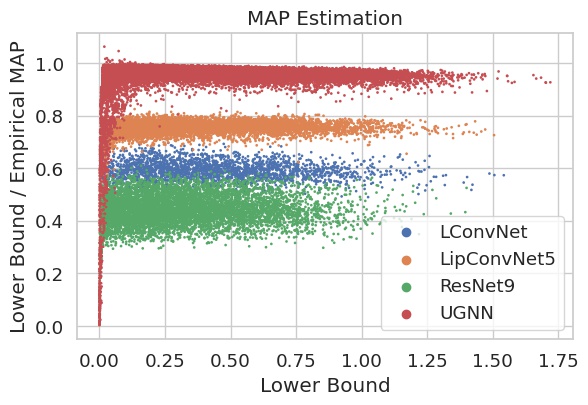}
    \subcaption{Raw Inputs (not normalized). UGNN with MaxMin and orth.}
  \end{subfigure}%\hfill
  \begin{subfigure}{\columnwidth}
    \includegraphics[width=\textwidth, keepaspectratio]{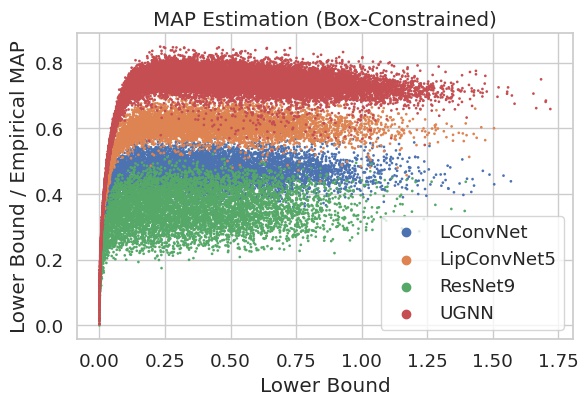}
    \subcaption{Raw Inputs (not normalized). UGNN with MaxMin and orth.}
  \end{subfigure}%
  \caption{MAP estimation among models without box constraints. For the empirical MAP computation, we used IP (a, c) for the Box-Unconstrained case and DDN for the constrained (b,d). Note that, in image (c), due to a failure in the IP algorithm convergence, few samples (less than $20$) reported a inconsistent MAP lower than expected.} 
   \label{fig:dist-multi}
\end{figure*}

\begin{figure}[ht]
    \centering
    \includegraphics[width=\columnwidth]{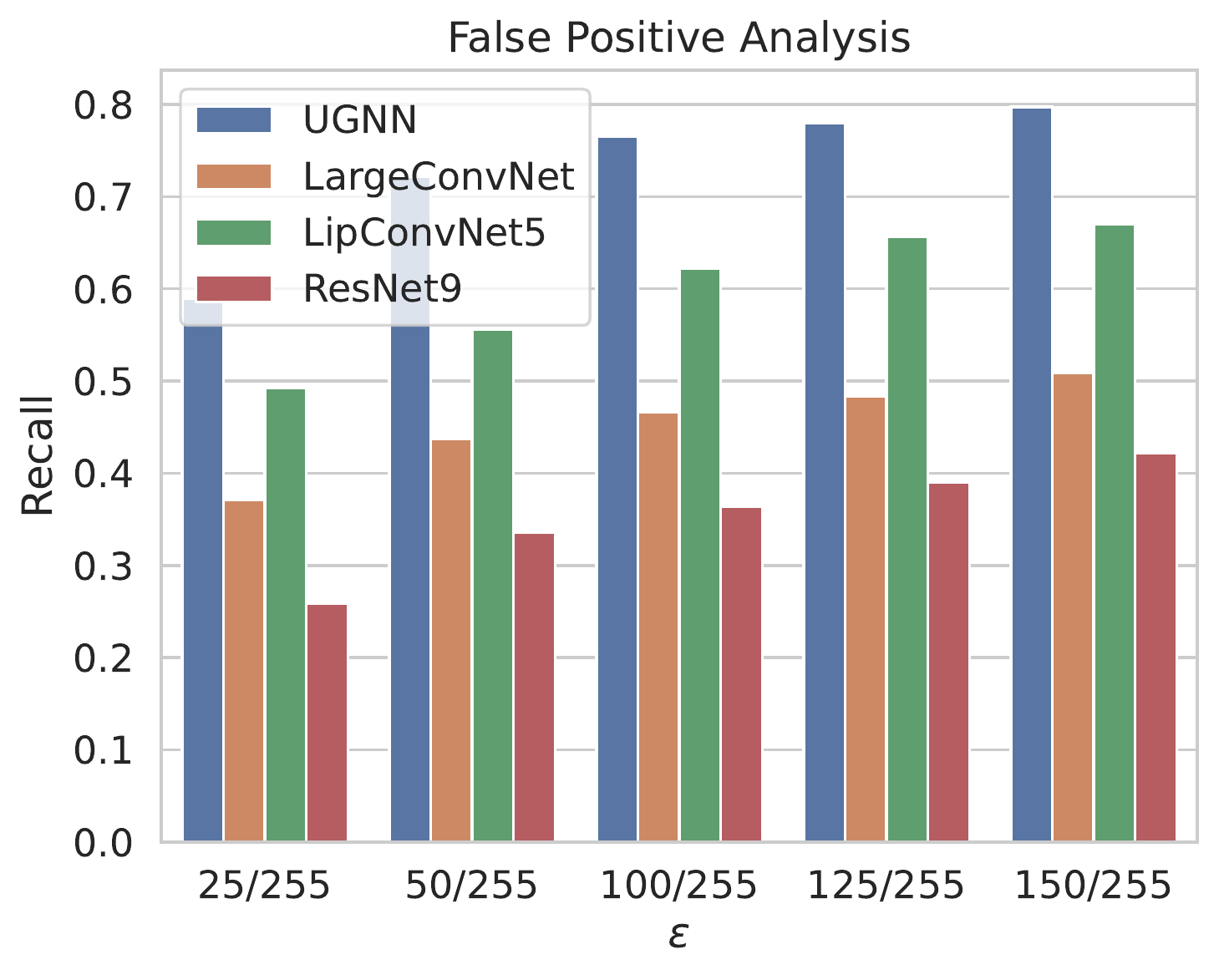}
    \caption{False-positive analysis}
    \label{fig:false-postive}
\end{figure}

Table~\ref{tab:further-map} and Figure~\ref{fig:dist-multi} show the MAP estimation through the lower bound provided by the tested models for different cases.
%------------------------
\begin{table}[ht]
    \centering
    \begin{tabular}{lccl}
\toprule
                  \textbf{Model} &  \textbf{LB/MAP} &  \textbf{\#N} & \textbf{B.C} \\
\midrule
                  ResNet9 (norm) & $.21\!\pm\!.042$ &          7900 &       \cmark \\
                   ResNet9 (raw) & $.34\!\pm\!.063$ &          6669 &       \cmark \\
             LargeConvNet (norm) & $.36\!\pm\!.036$ &          2148 &       \cmark \\
              LipConvNet5 (norm) & $.41\!\pm\!.060$ &          7838 &       \cmark \\
              LargeConvNet (raw) & $.46\!\pm\!.057$ &          7219 &       \cmark \\
               LipConvNet5 (raw) & $.58\!\pm\!.069$ &          6911 &       \cmark \\
  \textbf{UGNN+OPLU+updB (norm)} & ${\bf .67\!\pm\!.129}$ &          7101 &       \cmark \\
   \textbf{UGNN+Abs+updB (norm)} & ${\bf .67\!\pm\!.129}$ &          7220 &       \cmark \\
  \textbf{UGNN+OPLU+updU (norm)} & ${\bf .67\!\pm\!.131}$ &          7281 &       \cmark \\
   \textbf{UGNN+Abs+updU (norm)} & ${\bf .67\!\pm\!.128}$ &          7244 &       \cmark \\
\textbf{UGNN+MaxMin+updU (norm)} & ${\bf .67\!\pm\!.130}$ &          7311 &       \cmark \\
\textbf{UGNN+MaxMin+updB (norm)} & ${\bf .69\!\pm\!.109}$ &          4386 &       \cmark \\
   \textbf{UGNN+OPLU+updU (raw)} & ${\bf .70\!\pm\!.090}$ &          7125 &       \cmark \\
   \textbf{UGNN+OPLU+updB (raw)} & ${\bf .70\!\pm\!.093}$ &          7098 &       \cmark \\
 \textbf{UGNN+MaxMin+updB (raw)} & ${\bf .71\!\pm\!.087}$ &          7114 &       \cmark \\
 \textbf{UGNN+MaxMin+updU (raw)} & ${\bf .71\!\pm\!.088}$ &          7118 &       \cmark \\
    \textbf{UGNN+Abs+updB (raw)} & ${\bf .71\!\pm\!.090}$ &          6960 &       \cmark \\
    \textbf{UGNN+Abs+updU (raw)} & ${\bf .71\!\pm\!.092}$ &          6940 &       \cmark \\
    \midrule
                  ResNet9 (norm) & $.26\!\pm\!.036$ &          7904 &       \xmark \\
                   ResNet9 (raw) & $.44\!\pm\!.056$ &          6663 &       \xmark \\
             LargeConvNet (norm) & $.44\!\pm\!.027$ &          7933 &       \xmark \\
              LipConvNet5 (norm) & $.52\!\pm\!.031$ &          7840 &       \xmark \\
              LargeConvNet (raw) & $.58\!\pm\!.046$ &          2429 &       \xmark \\
               LipConvNet5 (raw) & $.74\!\pm\!.049$ &          6912 &       \xmark \\
   \textbf{UGNN+OPLU+updU (raw)} & ${\bf .93\!\pm\!.101}$ &          6755 &       \xmark \\
   \textbf{UGNN+OPLU+updB (raw)} & ${\bf .95\!\pm\!.063}$ &          7102 &       \xmark \\
 \textbf{UGNN+MaxMin+updU (raw)} & ${\bf .95\!\pm\!.061}$ &          7127 &       \xmark \\
 \textbf{UGNN+MaxMin+updB (raw)} & ${\bf .95\!\pm\!.056}$ &          7117 &       \xmark \\
    \textbf{UGNN+Abs+updB (raw)} & ${\bf .95\!\pm\!.058}$ &          6965 &       \xmark \\
    \textbf{UGNN+Abs+updU (raw)} & ${\bf .95\!\pm\!.058}$ &          6949 &       \xmark \\
  \textbf{UGNN+OPLU+updB (norm)} & ${\bf .96\!\pm\!.046}$ &          7215 &       \xmark \\
   \textbf{UGNN+Abs+updB (norm)} & ${\bf .96\!\pm\!.049}$ &          7228 &       \xmark \\
  \textbf{UGNN+OPLU+updU (norm)} & ${\bf .96\!\pm\!.047}$ &          7282 &       \xmark \\
\textbf{UGNN+MaxMin+updU (norm)} & ${\bf .96\!\pm\!.051}$ &          7316 &       \xmark \\
\textbf{UGNN+MaxMin+updB (norm)} & ${\bf .96\!\pm\!.044}$ &          7327 &       \xmark \\
   \textbf{UGNN+Abs+updU (norm)} & ${\bf .96\!\pm\!.039}$ &          7247 &       \xmark \\
\bottomrule
\end{tabular}

    \caption{Further tests of the evaluation of the estimation of the Minimal Adversarial Perturbation with many different configurations. This is a improved version of Table~\ref{tab:dist-table}}
    \label{tab:further-map}
\end{table}
%------------------------
%Table~\ref{tab:further-map} and Figure~\ref{fig:dist-multi} shows the MAP estimation through the lower bound analysis in different cases.
The bar-plot in Figure~\ref{fig:false-postive} contains, for each value of $\varepsilon$, the ratio of samples for which the classification is not $\varepsilon$-robust according to the lower-bound, but that feature a minimal adversarial perturbation larger than $\varepsilon$. The higher the bar, the fewer the practically $\varepsilon$-robust classifications discarded as not robust due to a lose lower bound. 
Let assume the following definitions, 
\begin{equation}
    TP_\varepsilon:=\{x\,:\, MAP(x)\le \varepsilon\,\land\,LB(x)\le\varepsilon\}
\end{equation}
\begin{equation}
    FP_\varepsilon:=\{x\,:\, MAP(x)> \varepsilon\,\land\,LB(x)\le\varepsilon\}
\end{equation}
\begin{equation}
    \label{eq:recall}
    r = \frac{\# TP_\varepsilon}{\# (TP_\varepsilon \cup FP_\varepsilon)}
\end{equation}
then, each bar of Figure~\ref{fig:false-postive} represents the value $r$ expressed by Equation~\eqref{eq:recall}.
\subsection{Further Robustness Evaluation}
 \begin{figure}[ht]
  \begin{subfigure}{\columnwidth}
  \includegraphics[width=\columnwidth, keepaspectratio]{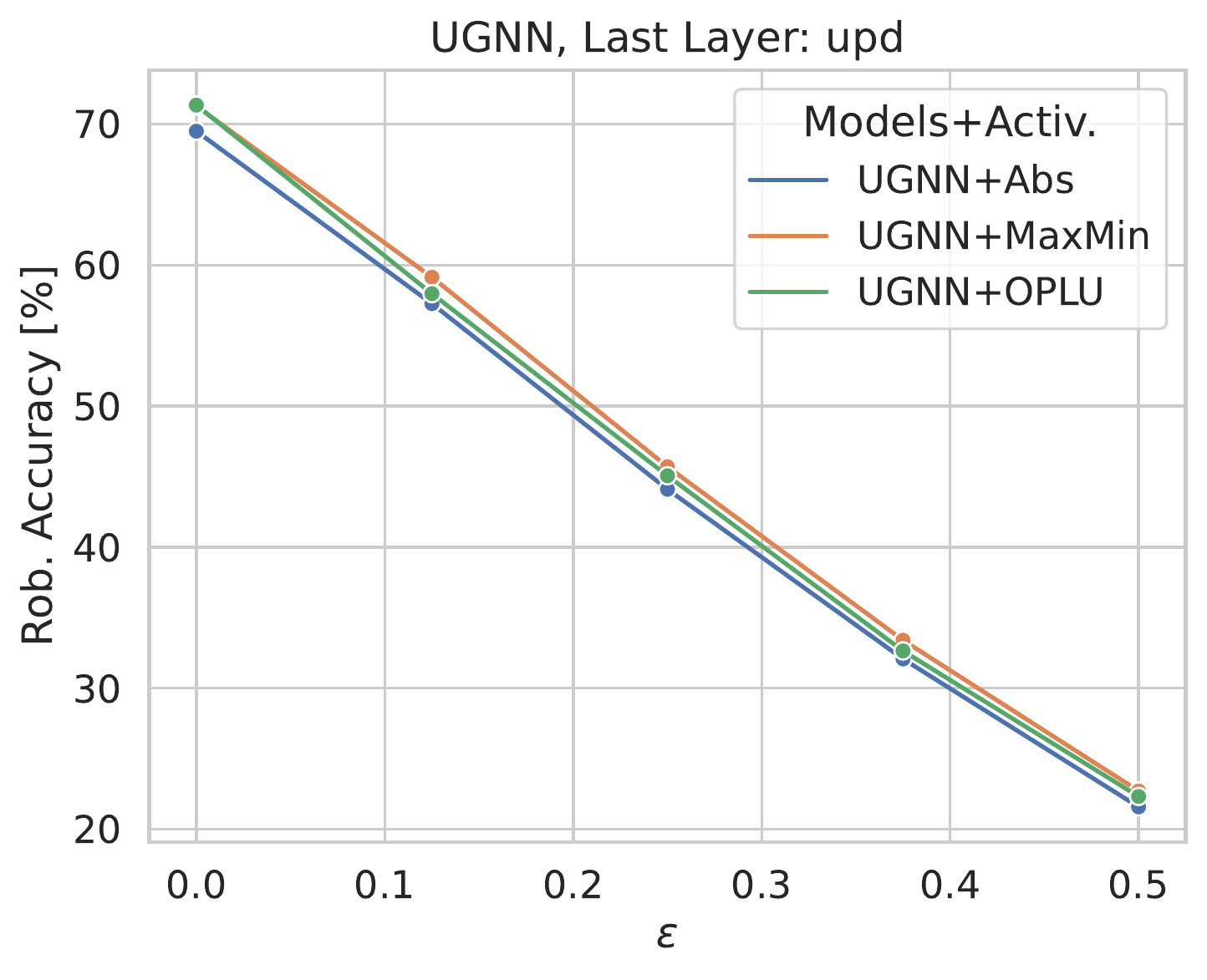}
  \caption{}
  \end{subfigure}
  %\hfill
  \begin{subfigure}{\columnwidth}
    \includegraphics[width=\columnwidth, keepaspectratio]{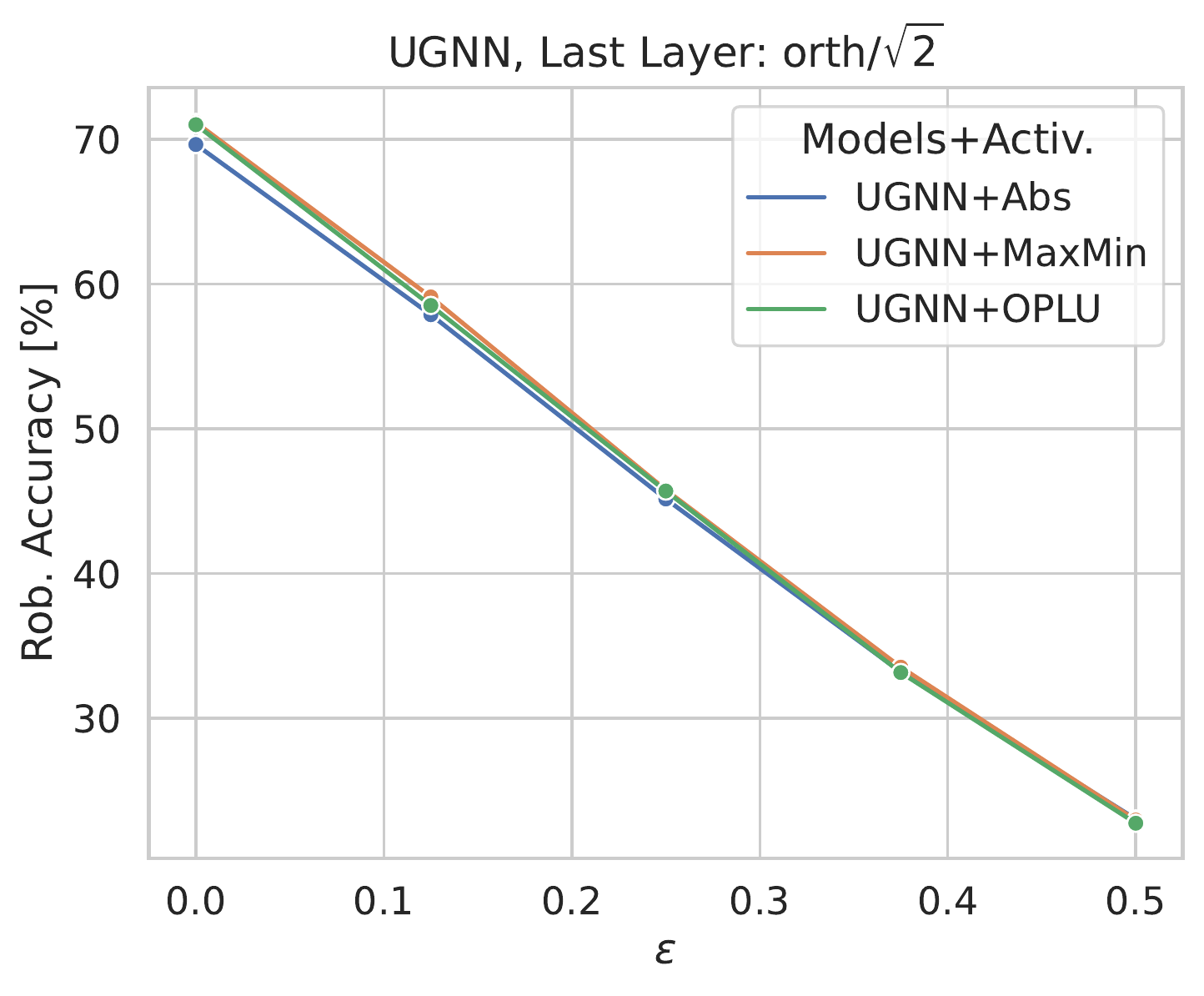}
    \caption{}
  \end{subfigure}
  \begin{subfigure}{\columnwidth}
    \includegraphics[width=\columnwidth]{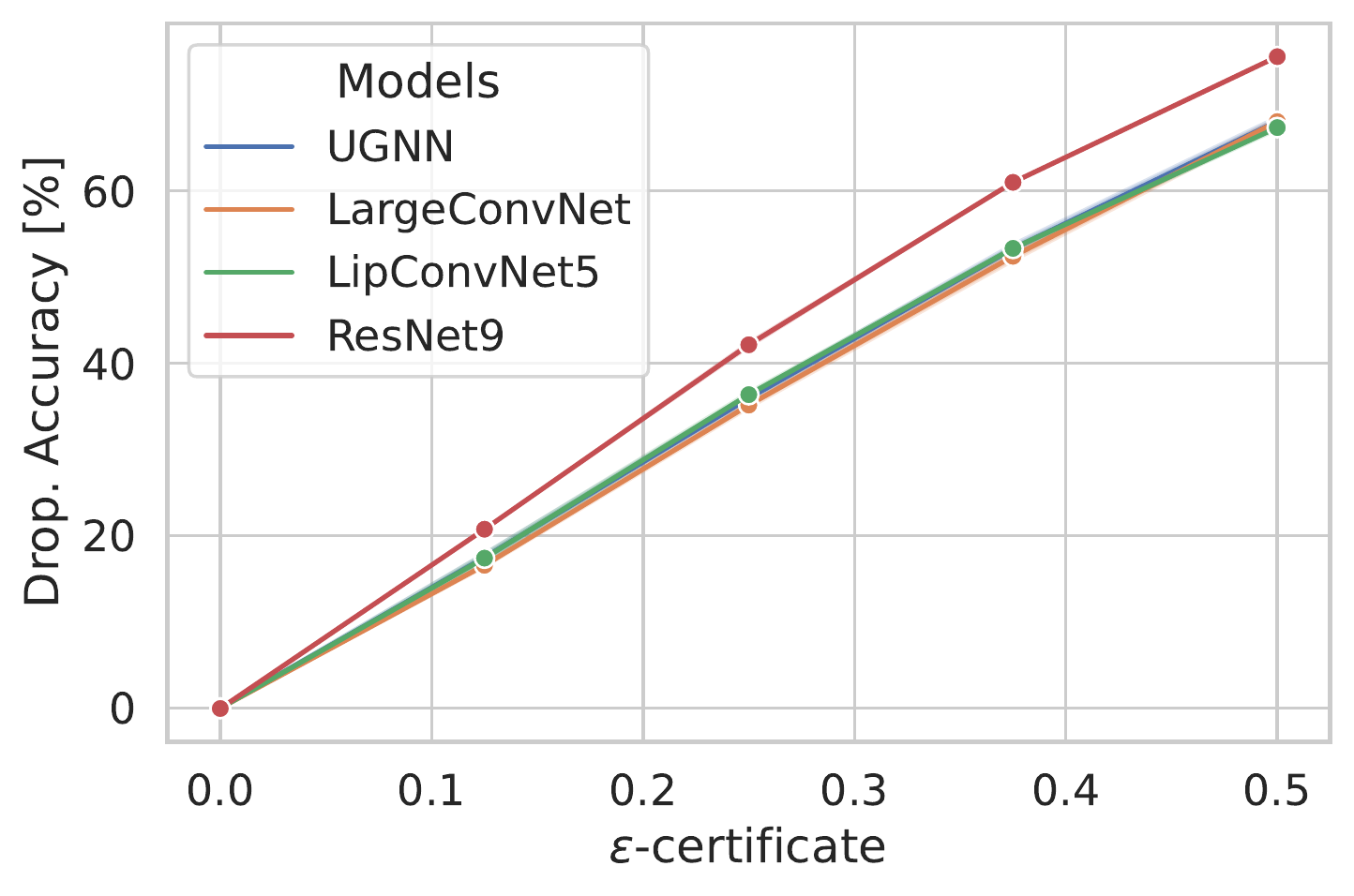}
    \caption{}
    %\label{fig:robust-drop}
  \end{subfigure}
  \caption{Accuracy of the $\varepsilon$-robust classifications for the UGNN with unbounded upd (a) and bounded upd (b). Figure (c) instead shows the relative drop of accuracy for the certifiable $\varepsilon$-robust classifications.}
  \label{fig:robust-lineplot}
\end{figure}
As it can be observed from Figure~\ref{fig:robust-lineplot}, for the case of 32x32 inputs we noted the same relative drop of accuracy of the other $1$-Lipschitz models proposed in previous work.%,~\ref{fig:robust-drop}.
%
% \begin{figure}[ht]
%     \centering
%     \includegraphics[width=\columnwidth]{img/robust_accuracy_drop.pdf}
%     \caption{Relative drop of accuracy for the certifiable $\varepsilon$-robust classifications.}
%     \label{fig:robust-drop}
% \end{figure}
%\clearpage
\onecolumn
\section{Architecture Code}
\begin{lstlisting}[language=Python, caption={UGNN implementation as \texttt{torch.nn.Module} used through all the experiments}]
class UGNN(nn.Module):
    def __init__(self, conv, linear, activation, nclasses = 10, in_ch=3, unitary_pair_diff=True):
        super().__init__()
        r'''
        The model takes input images of shapes in_ch x 2^k x 2^k and k >= 5
        '''
        assert in_ch in {1, 3}
        # dimensions
        self.in_ch = in_ch
        self.nclasses = nclasses
        self.activation = activation
        self.conv = conv
        self.linear = linear
        self.depth = 5
        self.feature_extraction = self.make_layers(kernel_size=3)
        self.flatten = nn.Flatten()
        self.fc = nn.Sequential(
            self.linear(in_ch*4**(self.depth), 1024), self.activation(),
            self.linear(1024, 512), self.activation(),
        )
        if unitary_pair_diff:
            self.last_fc = UnitaryPairDiff(512, nclasses)
        else:
            self.last_fc = nn.Sequential(
                self.linear(512, nclasses),
                Rescale())

    def _last_block(self):
        last_ch = self.in_ch*4**(self.depth-1)
        layer = nn.Sequential(
            self.conv(last_ch, last_ch, 3), self.activation(),
            self.conv(last_ch, last_ch, 3), self.activation(),
            nn.AdaptiveMaxPool2d((2, 2)),
            nn.PixelUnshuffle(downscale_factor=2)
        )
        return layer

    def _inner_block(self, channels, kernel_size):
        if channels % 2 != 0:
            activation = Abs
        else:
            activation = self.activation
        layer = nn.Sequential(
            self.conv(channels, channels, kernel_size), activation(),
            self.conv(channels, channels, kernel_size), activation(),
            nn.PixelUnshuffle(downscale_factor=2)
        )
        return layer

    def make_layers(self, kernel_size=3):
        layers = list()
        for idx in range(0, self.depth-1):
            layers.append(self._inner_block(self.in_ch*4**idx, kernel_size))
        layers.append(self._last_block())
        return nn.Sequential(*layers)

    def forward(self, x):
        aux = self.feature_extraction(x)
        aux = self.flatten(aux)
        aux = self.fc(aux)
        return self.last_fc(aux)
\end{lstlisting}

\begin{lstlisting}[language=Python, caption={Implementation of the parameterized unitary pair difference (updU) layer}]
STEPS = 3
def pairdiff_loss(weight: torch.Tensor) -> torch.Tensor:
    def _diff_matrix(m: int) -> torch.Tensor:
        if m < 2:
            raise ValueError
        if m == 2:
            return torch.Tensor([[1, -1]])
        ones = torch.ones(m-1, 1)
        eye = torch.eye(m-1)
        A_prev = _diff_matrix(m-1)
        zeros = torch.zeros(A_prev.shape[0], 1)
        A_up = torch.cat((ones, -eye), dim=1)
        A_low = torch.cat((zeros, A_prev), dim=1)
        return torch.cat((A_up, A_low), dim=0)
    T = _diff_matrix(weight.shape[0]).to(weight.device)
    return ((T@weight).pow(2).sum(1) - 1).pow(2).mean()

def unitary_pairdiff_proj(weight: nn.Parameter) -> nn.Parameter:
    from torch.optim import LBFGS
    optimizer = LBFGS([weight])

    def closure():
        optimizer.zero_grad()
        loss = pairdiff_loss(weight)
        loss.backward()
        return loss
    for _ in range(STEPS):
        optimizer.step(closure)
    return weight
    
class UnitaryPairDiff(nn.Linear):
    def forward(self, input: torch.Tensor) -> torch.Tensor:
        if self.train:
            self.proj_weight = unitary_pairdiff_proj(self.weight)
        if not hasattr(self, 'proj_weight'):
            with torch.no_grad():
                self.proj_weight = unitary_pairdiff_proj(
                    self.weight)
        return nn.functional.linear(input, self.proj_weight, self.bias)
\end{lstlisting}

\end{document}